\pdfoutput=1

\documentclass[11pt]{article}

\usepackage[final]{acl}

\usepackage{times}
\usepackage{latexsym}

\usepackage[T1]{fontenc}

\usepackage[utf8]{inputenc}

\usepackage{microtype}

\usepackage{inconsolata}

\usepackage{graphicx}
\usepackage{amsmath}
\usepackage{subfigure}
\usepackage{algorithm}
\usepackage{algpseudocode}
\usepackage{amssymb}
\usepackage{amsfonts}
\usepackage{multirow}
\usepackage{colortbl}
\title{Logit Separability-Driven Samples and Multiple Class-Related Words Selection for Advancing In-Context Learning}

\author{Zhu Zixiao\textsuperscript{1,3} \quad Zijian Feng\textsuperscript{1,3} \quad \textbf{Hanzhang Zhou\textsuperscript{1,3} \quad Junlang Qian\textsuperscript{2} \quad Kezhi Mao\textsuperscript{2}\thanks{Corresponding author}} \\
        \textsuperscript{1}Institute of Catastrophe Risk Management, Interdisciplinary Graduate Programme,\\ Nanyang Technological University, Singapore \quad \\ \textsuperscript{2}School of Electrical and Electronic Engineering, Nanyang Technological University, Singapore \quad \\ \textsuperscript{3}Future Resilient Systems Programme, Singapore-ETH Centre, CREATE campus, Singapore \\
        \texttt{\{zixiao001, junlang001, feng0119, hanzhang001\}@e.ntu.edu.sg}, \texttt{ekzmao@ntu.edu.sg}}

\begin{document}
                
\maketitle

\begin{abstract}
Effective organization of in-context learning (ICL) demonstrations is key to improving the quality of large language model (LLM) responses. To create better sample-label pairs that instruct LLM understanding, we introduce logit separability, a criterion to assess the clarity of both samples and class-related words at the logit level. This facilitates the optimization of sample and label selection, enhancing the precision of information provided in ICL demonstrations. Additionally, we find that incorporating multiple class-related words for each sample, rather than relying on a single class name, improves performance by offering a broader range of label information. Building on these insights, we propose LICL, a logit separability-based method that jointly organizes samples and integrates multiple class-related words into each sample-label pair. Evaluations across seven classification datasets show that this approach significantly improves ICL performance by providing clearer instructions and richer label information\footnote{Our code is available at \url{https://github.com/MidiyaZhu/MICL}}.

\end{abstract}

\section{Introduction}\label{sec:introduction}
In-context learning (ICL) enables large language models (LLMs) to perform new tasks using sample-label pairs as demonstrations, without the need for retraining or fine-tuning \citep{brown2020language}. However, the organization of these sample-label pairs is critical, as it can significantly affect ICL performance\footnote{In this paper, we follow \citet{wu2023self} to denote the selection and ranking of sample-label pairs as organization.} \citep{liu2022makes}. 

How do we assess the suitability of samples and labels for demonstration? In traditional machine learning, effective features are those that exhibit high discriminative power, enabling clear distinction between their respective classes \citep{fukunaga2013introduction}. Analogously, we introduce \textbf{logit separability}, a criterion to assess how well in-demonstration samples and label words differentiate between classes at the logit level. Logit separability captures two aspects: (1) how significantly a sample is predicted to its true label, where the correct label has substantially higher logit than others, and (2) how consistently a class-related word yields high logit values across samples of the same label. This measure provides insight into both the separability of samples through class-related words and the ability of these words to distinguish between samples of different classes. 

\begin{figure*}[t]
\centering  
\begin{minipage}[t]{0.6\textwidth}  
    \centering
    \subfigure[Samples' logit separability in ZSL]{
        \includegraphics[height=6cm, width=0.96\textwidth]{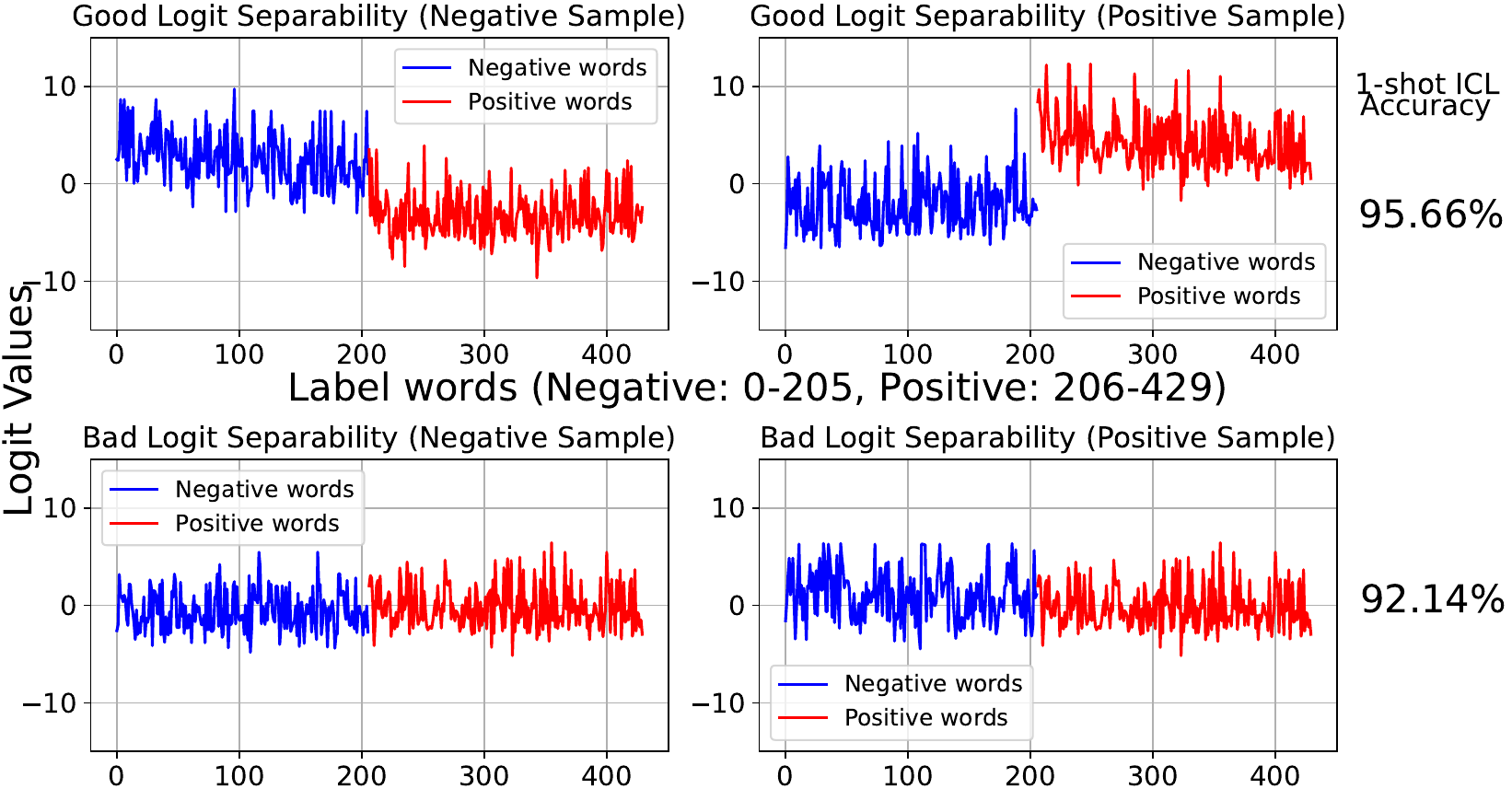}
        \label{effectiveness of semantics in sample}
    }
\end{minipage}%
\begin{minipage}[t]{0.4\textwidth}  
    \centering
    \subfigure[Class-related words output distribution in ZSL 
    ]{
        \includegraphics[height=3.5cm, width=\textwidth]{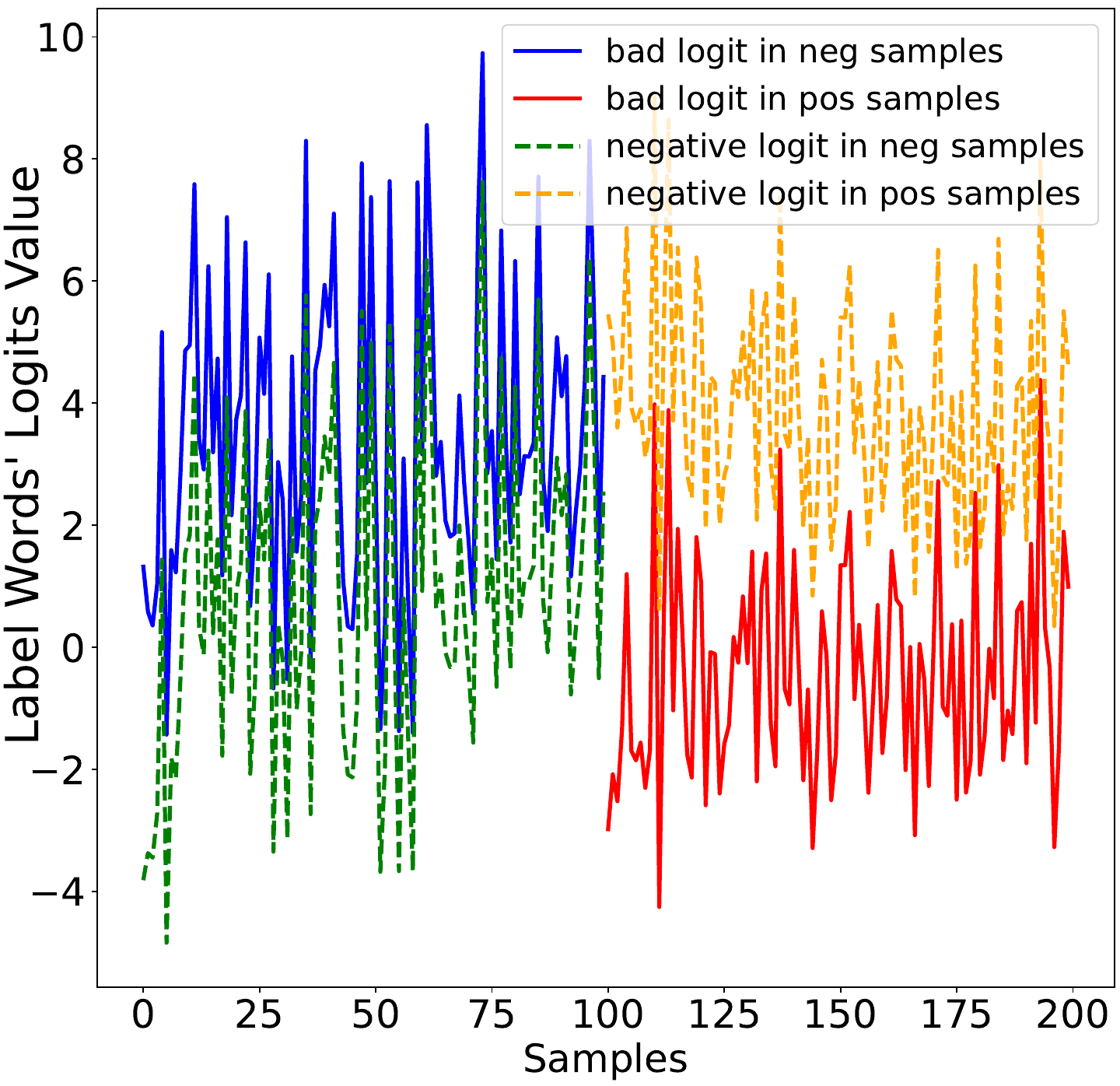}
        \label{distribution}
    }
    \subfigure[Multiple class-related words effectiveness in ICL]{
        \includegraphics[height=2.5cm, width=\textwidth]{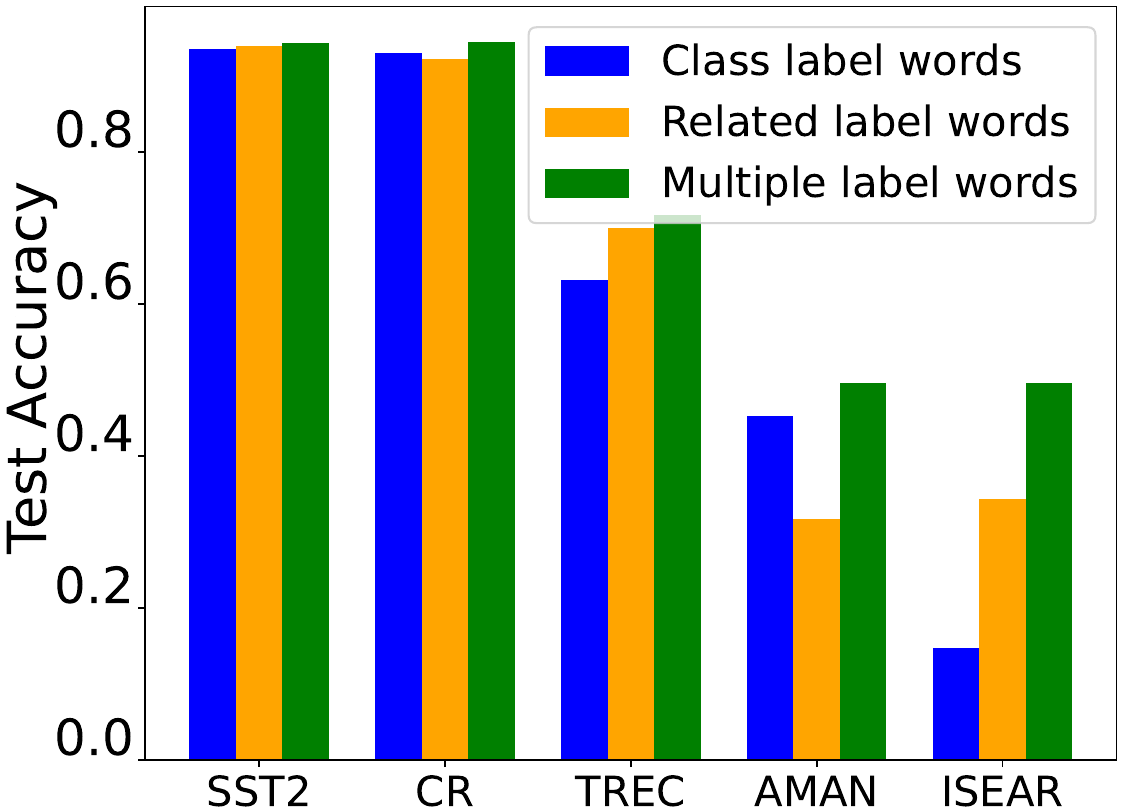}
        \label{mlabel}
    }
\end{minipage}
\caption{Exploration of Samples and class-related words on LLaMA2-7b. (a): Logit separability of SST-2 samples across class-related words under zero-shot learning (ZSL), showing varying degrees of separation due to different input samples. The 1-shot accuracy is demonstrated using the good or bad negative/positive samples, with class names as labels. (b): Logit values of various class-related words for 100 negative and 100 positive SST-2 samples under ZSL, showcasing the logit separability of class-related words across samples. (c): Accuracy comparison in 1-shot ICL using class names, single class-related words, and multiple class-related words (combining the two sets with spaces) as labels. Performance with multiple class-related words surpassed the other two sets. More experiments and analyses, including those on GPT2-xl, are in Appx.\ref{appendixa}.}
\label{label efficiency}
\end{figure*}

Building on this concept, we observe that not all samples or class-related words exhibit the same degree of logit separability. As shown in Fig.\ref{effectiveness of semantics in sample}, some samples produce strong separation in the logits of class-related words, assigning high logits to words matching the true label and low logits to others, while others show marginal differences between them. 
Additionally, certain class-related words consistently display higher logit in samples of their corresponding label compared to samples of other labels. As shown in Fig.\ref{distribution}, "bad" shows a clearer differentiation between positive and negative samples than the class name "negative".

Given the importance of providing clear label-instructed information in demonstrations, leveraging logit separability for selecting samples and labels might enhance ICL performance. As shown in Fig.\ref{effectiveness of semantics in sample}-accuracy, samples with strong logit separability provide more precise signals by aligning class-related words with the model's internal representations. Besides, class-related words with distinct logit separability can beat predefined class names in guiding the model toward accurate predictions (Appx.\ref{appendixa1}). Therefore, selecting both samples and class-related words via logit separability ensures clearer, more effective demonstrations, ultimately improving the performance of ICL.

Besides, the variability observed in logit values of class-related words across different samples shows that relying on a single class-related word may not adequately capture the full semantics associated with a class name. Unlike expanding candidates via a verbalizer in prediction mapping, directly inserting class-related words into demonstrations offers explicit instructions that guide LLM responses to potential candidates. Consequently, transitioning from single to multiple class-related words could enhance linguistic precision and deepen the semantic representation of class names, potentially leading to enhanced model performance. Fig.\ref{mlabel} supports that using multiple class-related words in sample-label pairs strengthens ICL performance.

Given samples and labels' critical roles in demonstration effectiveness for ICL, in this work, we propose a logit separability-based demonstration organization method. We organize demonstration samples based on their zero-shot logit separability across a pool of class-related words. In demonstration label selection, we propose a novel method that utilizes multiple class-related words instead of a single one. The selection, ordering, and quantity of class-related words are determined through sequential forward search, guided by logit feedback from the selected samples and their validation performance. Moreover, while most research investigates either the sample selection \citep{zhang2022active, hongjin2022selective, levy-etal-2023-diverse, lu-etal-2022-fantastically} or label selection \citep{yoo2022ground, milios2023context, gao2023benefits, peskine2023definitions} individually, we comprehensively address both aspects to enhance ICL quality.

We summarize our contributions as follows: 

\noindent1. We introduce \textbf{Logit Separability}, a criterion that evaluates how clearly a sample or label distinguishes between classes at the logit level, helping assess the suitability of samples and labels for demonstrations in ICL.
\\ \noindent2. We propose \textbf{LICL}, a \textbf{L}ogit separability-based demonstration organization method for \textbf{ICL}. We strategically select and order both samples and labels via the class-related words' logit separability in LLM’s output space, thereby improving the demonstration's instruction efficacy.
\\ \noindent3. We present a novel label instruction method for ICL that inserts multiple class-related words in sample-label pairs. This approach provides richer label information, improving breadth, reducing ambiguity, and enhancing ICL performance.

\section{Related Work}

\noindent\textbf{Demonstration organization in ICL} To improve ICL performance with better demonstration organization, some studies use pre-trained models like S-BERT \cite{liu2022makes} or BM25 \cite{hongjin2022selective, levy-etal-2023-diverse} to select and rank demonstrations. While these unsupervised methods have advantages, they may cause inconsistencies in knowledge transfer. Other approaches organize demonstration based on the LM's output distribution. Some methods take part of the training set as validation to enable supervised learning methods for demonstration organization \citep{chang2023data, zhang2022active, wang2024large}. However, this will shrink the pool of candidates, risking sub-optimal selection. Additionally, some methods use the LM's output, like label confidence, to organize demonstrations under the full training set \cite{rubin2022learning, wu2023self, li2023finding}. In the above works, the class names are preassumed to be the label word when creating the demonstration sample-label pairs. 

\noindent\textbf{Label matters in ICL} The significance of demonstration labels in ICL has been debated. \citet{yoo2022ground} demonstrates a positive relationship between ICL performance and accurate sample-label mapping. \citet{li2023finding} reveals that using the same sample but different class-related words as labels in the demonstrations can result in very different ICL performance. 
\citet{wang2023label} suggests that labels derive semantic representations from demonstrations for use in deep layers to make final predictions. \citet{yu2024large} further shows that these demonstration features are integrated into corresponding labels, with each in-context head extracting features specific to these labels. \citet{milios2023context} uses different class-related words in augmented samples to enhance ICL performance, indicating the potential for enhanced demonstration effectiveness when multiple class-related words are used. However, this approach significantly increases the required token length and running time in augmentation settings.

\section{Method}
\begin{figure*}[t] 
\centering
\includegraphics[height=6cm, width=0.9\textwidth]{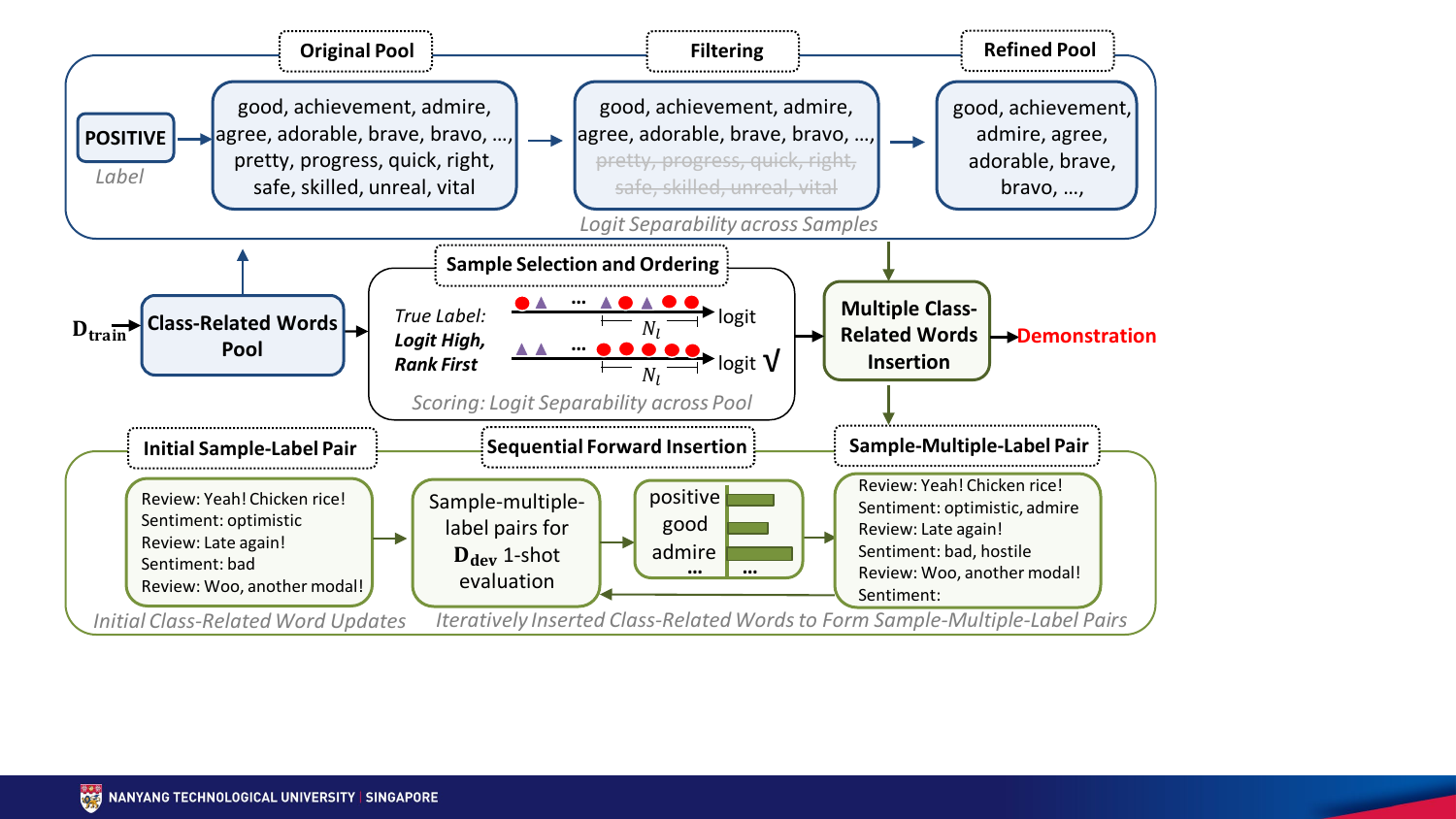}
\caption{Overall architecture of LICL: The top part shows the pool refinement, with sample organization based on logit separability across the refined pool. The bottom part presents multiple class-related word insertion via sequential forward search, starting from an initial sample-label pair to form a sample-multiple-label pair.}
\label{The model architecture of the proposed method.}
\end{figure*}
In this section, we introduce our proposed method \textbf{LICL}, which comprises logit separability-based demonstrated sample selection and ordering, and corresponding multiple class-related words insertion, as shown in Fig.\ref{The model architecture of the proposed method.}. 
\subsection{Problem Statement\label{sec3}}
Give a large language model $M$, label space (class names) $L$, a class-related word pool $P$ with class-related words (includes class names), test sample $x_{test}$ and demonstrations $\left\{ x_{i},y_{i}\right\}_{i=1}^L $\footnote{Our work focus on 1-shot ICL with label-balanced demonstrations, so labels in demonstration are from the $1$-th to the $L$-th label. Discussion of label balance can be found in Appx.\ref{sec:labelbalance}}.
The zero-shot classification of $x_{test}$ can be based on the logit of class names as $\arg\max_{y \in L} p_M(y \mid x_{\text{test}})$ or based on the logit of all class-related words in the pool as $\arg\max_{y \in P} p_M(y \mid x_{\text{test}})$. Similarly, the 1-shot prediction of $x_{test}$ is $\arg\max_{y \in L} p_M(y \mid x_1 \oplus y_1, \cdots, x_L \oplus y_L \oplus x_{\text{test}})$ or $\arg\max_{y \in P} p_M(y \mid x_1 \oplus y_1, \cdots, x_L \oplus y_L \oplus x_{\text{test}})$. $\oplus$ is the concatenation operation, and function $p_M(\cdot)$ returns the logit of words in $M$'s vocabulary.

\subsection{Logit Separability-Based Sample Selection and Ordering\label{sampleselect}}

A well-chosen sample should exhibit strong logit separability, producing higher logits for class-related words that correspond to the sample's label, with these words ranked as highly as possible among all words in the pool $P$ based on logit values. This ensures that the selected samples are aligned with multiple correlated class-related words while minimizing alignment with uncorrelated words, enhancing the sample's class-related representativeness and improving the effectiveness of ICL demonstrations.

\noindent\textbf{Pool Refinement} The initial pool $P$ is constructed from existing works \cite{hu2022knowledgeable,zhu2024edentail} with details in Appx.\ref{appendixdatasets}. However, not all words in $P$ exhibit correct separable features within both the dataset and the LLM, risking the introduction of incorrect class-related features. To refine $P$ and maintain the quality of the logit separability measurement, we propose a filtering method to generate a refined pool, $P_r$, by selecting class-related words based on their logit separability and correlation using point-biserial testing \cite{kornbrot2014point} across dataset samples. Details of this refinement are in Appx.\ref{sec: labelwordfilter}, with analyses in Appx.\ref{sec:filtering}.

With the refined pool $P_r$ in place, for each zero-shot learning sample, the LLM generates logit values for all words in $P_r$. We first remove samples where the class-related word with the highest logit corresponds to an incorrect label. Next, we score samples for selection. Since predictions only rely on the word with the highest logit, scoring based on all words introduces noise—particularly in multi-class tasks—because the pool contains far more words linked to incorrect labels than to the correct label. To address this, we focus exclusively on the true label pool, using two methods—Top-Logit Summation and Rank-Weighted Counting. These methods ensure that selected samples align with logit separability across relevant labels, yielding higher logits for class-related words matching the sample’s label and ranking them among the top in $P_r$. Let $N_l$ be the number of words in $P_r$ for class $l$, and $P_{r}^{l}$ be the pool of class-related words for class $l$.

\noindent\textbf{Top-Logit Summation}
For each training sample $t_{j}$ with label $l_{t_{j}}$, we compute its score by summing the logits of the top $N_l$ words from $P_r$, sorted in descending order by their logits $b_{w_{i}}$ (Eq.\ref{eq:score1}). Ideally, these top $N_l$ words are all class-related words corresponding to $l_{t_{j}}$, indicating strong logit separability. Therefore, only the top $N_l$ words $w_{i}$, which belong to the sample's label $l_{t_{j}}$ (i.e., in $P_{r}^{l_{t_{j}}}$), are included in the sum, ensuring that the score reflects only the contribution from the correct class-related words.

\begin{equation} 
\text{score}_{t_j} = \sum_{i=0}^{N_l -1} b_{w_i},\text{ if } w_i \in P_{r}^{l_{t_j}}
\label{eq:score1}
\end{equation}

\noindent\textbf{Rank-Weighted Counting}
When the logit values of class-related words are negative, summing logit becomes less effective. To address this, we focus on the ranks of the class-related words rather than their logit values to capture logit separability when logit is negative. Each training sample $t_{j}$ with label $l_{t_{j}}$, is evaluated by assigning linear weights to its corresponding class-related words based on their ranks within the top $N_{l}$ words from $P_r$, sorted by their logits $b_{w_{i}}$ (Eq.\ref{eq:score2}). Higher-ranked words receive higher weights, emphasizing their prominence.

\begin{equation} 
\text{score}_{t_j} = \sum_{i=0}^{N_l -1} \frac{2(N_l - i)}{(N_l +1)N_l}  \text{, if } w_i \in P_{r}^{l_{t_j}}
\label{eq:score2}
\end{equation}

In k-shot ICL, for each class $l$, we select the top-k highest-scoring training samples labeled as $l$. We then construct the demonstration sequence by interleaving these samples across classes: first including the highest-scoring sample from each class, followed by the second-highest, and so on. This ordering ensures that the most logit-separable and representative samples appear early while maintaining balanced class representation, thereby reducing potential bias and enhancing the effectiveness of the demonstrations.

\subsection{Multiple Class-Related Words Insertion}
As analyzed in Sec.\ref{sec:introduction}, relying solely on class names in the sample-label pair may not provide sufficient semantic information for the labels. To address this, we aim to enrich the semantic diversity of label prompts by employing multiple class-related words, proposing the use of a sequence of class-related words to create a sample-multiple-label pair, thereby enhancing the effectiveness of demonstration labels. 

After confirming that the selected samples respond well to the multiple class-related words in $P_r$, our method sequentially inserts these words into a single sample-label pair to form a sample-multiple-label pair, which is more computationally and memory efficient than using multiple augmented sample-label pairs in ICL instruction.

The word insertion process is based on sequential forward search, where the sample-multiple-label pairs are evaluated as 1-shot demonstrations on the validation set. New class-related words are inserted based on the output logit values. Intuitively, since no demonstration is in place, the labels in the initial sample-label pairs typically use predefined class names. However, we find that predefined class names are not always optimal, as they might exhibit lower logit in zero-shot learning compared to other class-related words. Consequently, we construct the initial sample-label pairs by replacing weak logit response class names with class-related words that yield higher logit values within the selected sample set.

\noindent\textbf{Initial Class-Related Word Updates}
If a class name $l$ ranks below the top $1$ based on its logit score in zero-shot learning across the selected samples from $P_r$, we replace it with the word that has the highest logit value within the selected sample set. This joint process of assigning initial class-related words from the selected samples ensures that the resulting sample-label pairs are better tailored for the LLM's logit feedback.

\noindent\textbf{Sequential Forward Insertion:} After obtaining the initial sample-label pairs, we select and order class-related words to form sample-multiple-label pairs. For each label $l \in L$, we insert the word $w_{i}^{l}$ with the highest average logit value across all samples in $D_{dev}^{l}$ under ICL as Eq.\ref{eq:multi}:  
\begin{equation} 
w_{i}^{l} = \arg\max_{w \in P_{r}^{l}} \frac{1}{|D_{dev}^{l}|}\sum_{x \in D_{dev}^{l}}b_{w}^{x}
\label{eq:multi}
\end{equation}
where $D_{dev}$ is the validation set derived from the training set with the demonstration samples removed, $D_{dev}^{l}$ is the subset of $D_{dev}$ containing all samples of class $l$, and $b_{w}^{x}$ is the logit value of word $w$ for validation sample $x$. After each insertion, the selected word is removed from $P_{r}^{l}$ to prevent repeated insertions.

 This insertion is iterative, with the selected word updating sample-multiple-label pairs in each round. The iteration continues until no more candidates remain in $P_{r}^{l}$, or until further insertions reduce validation set performance. Examples of prompts for zero-shot, 1-shot, and 1-shot multiple class-related word learning are provided in Appx.\ref{appendixprompts}.

\section{Experiments}

In this section, we examine the capacities of our method in ICL from the following perspectives: (1) 1-shot ICL classification comparison between baseline models and LICL (Sec.\ref{sec:mainresult}); (2) 5-shot ICL (Sec.\ref{sec:5shot}); (3) Ablation studies on sample ordering (Sec.\ref{sec:permutation}) and multiple class-related words selection and ordering (Sec.\ref{sec:insertaba})
; (4) Scalability to larger LLMs (Sec.\ref{sec:lagerllm}); (5) General applicability of selected multiple class-related words (Sec.\ref{sec: cross}); and (6) Further analyses (Sec.\ref{sec:futherana}).

\subsection{Setups}
\noindent\textbf{Datasets} 
Seven datasets are evaluated, including: SST-2 \citep{sst2}, CR \citep{cr}, IMDB \citep{maas2011learning}, ISEAR \citep{scherer1994evidence}, AMAN \citep{aman2008using}, TREC-6 \citep{trec}, and AGNews \citep{zhang2015character}. The templates are provided by \citet{wang2023label}. The initial class-related word pools are drawn from \citet{hu2022knowledgeable} and \citet{zhu2024edentail}. Detailed information on the datasets, templates, and pool is given in Appx.\ref{appendixdatasets}.

\noindent\textbf{Experiment Settings} We conduct few-shot learning experiments using LLaMA2-7b 
 \cite{touvron2023llama} and GPT2-xl \cite{radford2019language} to test the effectiveness of our methods. 

For the baseline models, including \textbf{vanilla LLaMA2-7b}, \textbf{vanilla GPT2-xl}, \textbf{Topk} \cite{liu2022makes}, \textbf{SelfICL} \cite{wu2023self}, and \textbf{DataICL} \cite{chang2023data}, we adopt their proposed demonstration samples and pair them with class name and our multiple class-related words, respectively, to create sample-label pairs. Detailed descriptions of the baseline models, including comparing their methods with ours and an overview of their experimental settings, are in Appx.\ref{appendixsettings}.

For each model, the sequential class-related word forward insertion is evaluated on the validation set, which is the full training set excluding the samples selected for demonstrations. In Vanilla, TopK, and SelfICL, the full training set is split into a reduced training set and a validation set in an 8:2 ratio as demonstration samples vary with test samples. The demonstration samples are selected based on the reduced training set, while the multiple class-related word insertion is based on the validation set. The ICL performance is evaluated on the same test set for all experiments. 

\subsection{Main Results\label{sec:mainresult}}
We present the results of each model and LICL (ours) under two settings: with and without multiple class-related word insertion in 1-shot ICL. The outcomes, summarized in Table \ref{LICLoverperformance}, lead to the following conclusions:

\begin{table*}[t]
\tiny
\centering
\resizebox{\linewidth}{!}{
\begin{tabular}{rcccccccc}
\hline
\multicolumn{1}{c}{}                                      & SST2           & CR             & IMDB           & TREC           & AMAN           & ISEAR          & AGNews         & Avg.     \\ \hline
\multicolumn{1}{l}{\textit{LLaMA2-7b}}                    &                &                &                &                &                &                &                &                \\
\multicolumn{1}{l}{\textbf{vanilla-LLaMA2-7b}}            & 93.06          & 93.24          & 94.81          & 68.20          & 53.08          & 70.91          & 81.78          & 79.30          \\
+MLabels\_CN                                             & 93.74$\uparrow$          & 94.41$\uparrow$          & 95.89$\uparrow$          & 69.88$\uparrow$          & 59.75$\uparrow$          & 71.48$\uparrow$          & 83.36$\uparrow$          & 81.22          \\
+MLabels\_LW                                            & 93.74          & 94.41          & 95.89          & 72.88$\uparrow$          & 59.05          & 71.03          & 83.88$\uparrow$          & 81.55          \\ \hline
\multicolumn{1}{l}{\textbf{TopK}}                         & 92.37          & 92.82          & 94.29          & 77.80          & 51.38          & 64.32          & 79.09          & 78.87          \\
+MLabels\_CN                                             & 93.52$\uparrow$          & 93.55$\uparrow$          & 94.69$\uparrow$          & 83.20$\uparrow$          & 58.15$\uparrow$          & 67.38$\uparrow$          & 79.59$\uparrow$          & 81.44          \\
+MLabels\_LW                                            & 93.52          & 93.55          & 94.69          & 83.20          & 58.65$\uparrow$          & 66.98          & 79.59          & 81.45          \\ \hline
\multicolumn{1}{l}{\textbf{SelfICL}}                      & 91.71          & 93.35          & 94.69          & 76.00          & 53.63          & 65.18          & 81.46          & 79.36          \\
+MLabels\_CN                                             & 92.53$\uparrow$          & 92.82$\uparrow$          & 95.10$\uparrow$          & 79.00$\uparrow$          & 58.65$\uparrow$          & 69.10$\uparrow$          & 82.01$\uparrow$          & 81.39          \\
+MLabels\_LW                                            & 92.53          & 92.82          & 95.10          & \textbf{82.80}$\uparrow$ & 58.65          & 69.30$\uparrow$          & 82.01          & 81.96          \\ \hline
\multicolumn{1}{l}{\textbf{DataICL}}                      & 94.51          & 89.89          & 94.60          & 71.80          & 53.88          & 70.17          & 83.45          & 79.76          \\
+MLabels\_CN                                             & 95.28$\uparrow$          & 92.55$\uparrow$          & 94.70$\uparrow$          & 74.40$\uparrow$          & 55.64$\uparrow$          & 71.30$\uparrow$          & 84.83$\uparrow$          & 81.40          \\
+MLabels\_LW                                            & 95.28          & 92.55          & 94.70          & 78.80$\uparrow$          & 54.89          & 71.10          & 85.45$\uparrow$          & 81.95          \\ \hline
\multicolumn{1}{l}{\cellcolor[HTML]{ECF4FF}\textbf{LICL}} & 95.39\textsuperscript{1}          & 94.41\textsuperscript{1}          & 95.40\textsuperscript{1}          & 78.40\textsuperscript{1}          & 59.90\textsuperscript{2}          & 72.49\textsuperscript{2}          & 84.11\textsuperscript{2}          & 82.87          \\
\cellcolor[HTML]{ECF4FF}+MLabels\_CN                     & 95.97\textsuperscript{1}$\uparrow$          & 95.15\textsuperscript{1}$\uparrow$          & 95.60\textsuperscript{1}$\uparrow$          & 79.80\textsuperscript{1}$\uparrow$          & 65.16\textsuperscript{2}$\uparrow$          & \textbf{73.55}\textsuperscript{2}$\uparrow$ & 86.55\textsuperscript{2}$\uparrow$          & 84.54          \\
\cellcolor[HTML]{ECF4FF}+MLabels\_LW                    & \textbf{95.97}\textsuperscript{1} & \textbf{95.15}\textsuperscript{1} & \textbf{95.60}\textsuperscript{1} & 80.60\textsuperscript{1}$\uparrow$          & \textbf{69.40}\textsuperscript{2}$\uparrow$ & 73.09\textsuperscript{2}          & \textbf{86.58}\textsuperscript{2}$\uparrow$ & \textbf{85.20} \\ \hline
\multicolumn{1}{l}{\textit{GPT2-xl}}                      &                &                &                &                &                &                &                &                \\
\multicolumn{1}{l}{\textbf{vanilla-GPT2-xl}}              & 71.74          & 64.26          & 67.00          & 46.84          & 29.97          & 39.00          & 55.24          & 53.44          \\
+MLabels\_CN                                             & 85.63$\uparrow$          & 67.07$\uparrow$ & 68.23$\uparrow$          & 53.60$\uparrow$          & 39.35$\uparrow$          & 50.01$\uparrow$          & 58.44$\uparrow$          & 60.33          \\
+MLabels\_LW                                            & 85.63          & 67.07          & 70.13$\uparrow$          & 54.76$\uparrow$          & 39.30          & 50.56$\uparrow$          & 58.24         & 60.81          \\ \hline
\multicolumn{1}{l}{\textbf{TopK}}                         & 69.41          & 65.69          & 63.36          & 56.20          & 32.83          & 44.12          & 54.33          & 55.13          \\
+MLabels\_CN                                             & 84.51$\uparrow$          & 66.22$\uparrow$          & 65.47$\uparrow$          & 60.80$\uparrow$          & 40.85$\uparrow$          & 54.75$\uparrow$          & 55.24$\uparrow$          & 61.12          \\
+MLabels\_LW                                            & 84.51          & 66.22          & 67.07$\uparrow$          & 61.80$\uparrow$          & 40.60          & 54.88$\uparrow$          & 55.24          & 61.47          \\ \hline
\multicolumn{1}{l}{\textbf{SelfICL}}                      & 70.07          & 64.89          & 60.96          & 56.00          & 32.58          & 44.98          & 54.50          & 54.85          \\
+MLabels\_CN                                             & 83.80$\uparrow$          & 65.43$\uparrow$       & 61.46$\uparrow$          & 64.20$\uparrow$          & 43.36$\uparrow$          & 54.82$\uparrow$          & 56.42$\uparrow$          & 61.36         \\
+MLabels\_LW                                            & 83.80          & 65.43          & 62.16$\uparrow$          & 65.40$\uparrow$         & 43.11          & 55.08$\uparrow$          & 56.42          & 61.63          \\ \hline
\multicolumn{1}{l}{\textbf{DataICL}}                      & 83.47          & 63.83          & 64.80          & 57.20          & 35.34          & 35.28          & 43.36          & 54.75          \\
+MLabels\_CN                                             & 84.84$\uparrow$  & 65.74$\uparrow$    & 69.80$\uparrow$   & 58.20$\uparrow$  & 36.84$\uparrow$  & 48.04$\uparrow$   & 51.14$\uparrow$     & 59.23         \\
+MLabels\_LW                                            & 84.84          & 65.74          & 69.20          & 65.20$\uparrow$          & 36.84          & 48.04          & 51.14          & 60.14          \\ \hline
\multicolumn{1}{l}{\cellcolor[HTML]{ECF4FF}\textbf{LICL}} & 85.17\textsuperscript{2}          & 64.89\textsuperscript{1}          & 71.50\textsuperscript{1}          & 70.00\textsuperscript{2}            & 47.87\textsuperscript{2}            & 48.64\textsuperscript{2}            & 78.92\textsuperscript{2}            & 66.71          \\
\cellcolor[HTML]{ECF4FF}+MLabels\_CN                     & 91.65\textsuperscript{2}$\uparrow$   & \textbf{69.95}\textsuperscript{1}$\uparrow$ & 73.40\textsuperscript{1}$\uparrow$  & 70.40\textsuperscript{2}$\uparrow$ & 49.62\textsuperscript{2}$\uparrow$          & \textbf{59.73}\textsuperscript{2}$\uparrow$ & 79.08\textsuperscript{2}$\uparrow$          & \textbf{70.55} \\
\cellcolor[HTML]{ECF4FF}+MLabels\_LW                    & \textbf{91.65}\textsuperscript{2}   & \textbf{69.95}\textsuperscript{1}          & \textbf{73.40}\textsuperscript{1} & \textbf{70.40}\textsuperscript{2}            & \textbf{49.87}\textsuperscript{2}$\uparrow$ & 58.74\textsuperscript{2}            & \textbf{79.49}\textsuperscript{2}$\uparrow$ & 70.50          \\ \hline
\end{tabular}
}
\caption{1-Shot ICL Experimental Results: Model Name (e.g., TopK, LICL) is demonstrated on initial sample-label pairs while `+MLabels' is on sample-multiple-label pairs. `+MLabels\_CN' is predicting based on the maximum logit over class names, and `+MLabels\_LW' is based on the maximum logit over inserted class-related words. The best accuracy results (\%) are marked in bold. Marker \textsuperscript{1} indicates results given under Eq.\ref{eq:score1}, while marker \textsuperscript{2} indicates results given under Eq.\ref{eq:score2}. "$\uparrow$" in `+MLabels\_CN' signifies an increase in performance compared to the original method, while in `+MLabels\_LW', it signifies an increase in performance compared to `+MLabels\_CN'. }
\label{LICLoverperformance}
\end{table*}
\normalsize 

\noindent\textbf{LICL Outperforms Baselines:} The proposed LICL consistently outperforms baseline methods across all datasets under the initial single-label pair conditions. It achieves an average accuracy improvement of 3.11\% in LLaMA2-7b and 11.58\% in GPT2-xl. The gains are especially significant in multi-class classification tasks, with notable improvements of 6.02\% (AMAN, LLaMA2-7b), 12.80\% (TREC, GPT2-xl), 12.53\% (AMAN, GPT2-xl), and 23.68\% (AGNews, GPT2-xl).

\noindent\textbf{Multiple Class-Related Words Enhanced:} Inserting multiple class-related words (`+MLabels') significantly improves performance across all models. This strategy yields an average baselines' accuracy increase of 2.39\% in LLaMA2-7b, reaching up to 2.6\% in SelfICL, and an average increase of 5.94\% in GPT2-xl, with a maximum improvement of 7.37\% in vanilla GPT2-xl. Compared to configurations that employ a single class-related word per sample-label pair, LICL achieves notable accuracy gains, with increases of 2.33\% in LLaMA2-7b (peaking at 9.6\% in AMAN) and 3.84\% in GPT2-xl (peaking at 11.09\% in ISEAR). These improvements demonstrate significant enhancements over the initial single-label setup, underscoring the efficacy of the multiple class-related word approach.

\begin{figure*}[htbp]
\centering  
\subfigure[SST-2]{
\label{sst2}
\includegraphics[height=2.6cm, width=0.32\textwidth]{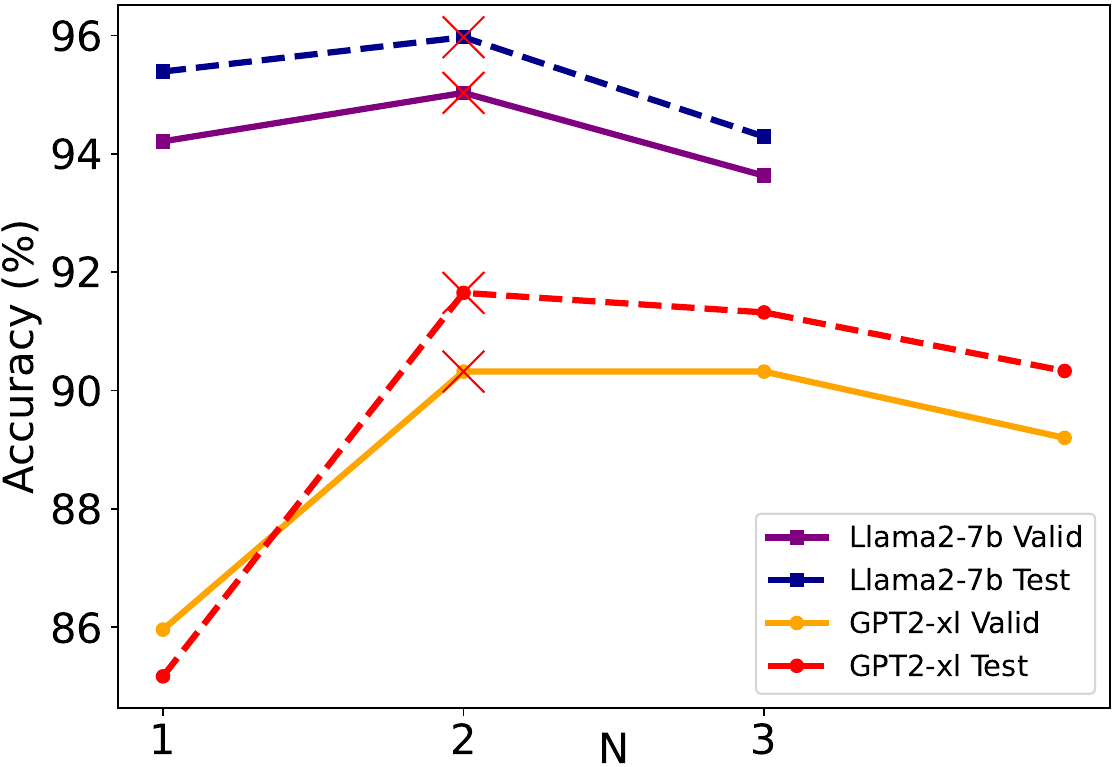}}
\subfigure[TREC]{
\label{trec}
\includegraphics[height=2.6cm, width=0.32\textwidth]{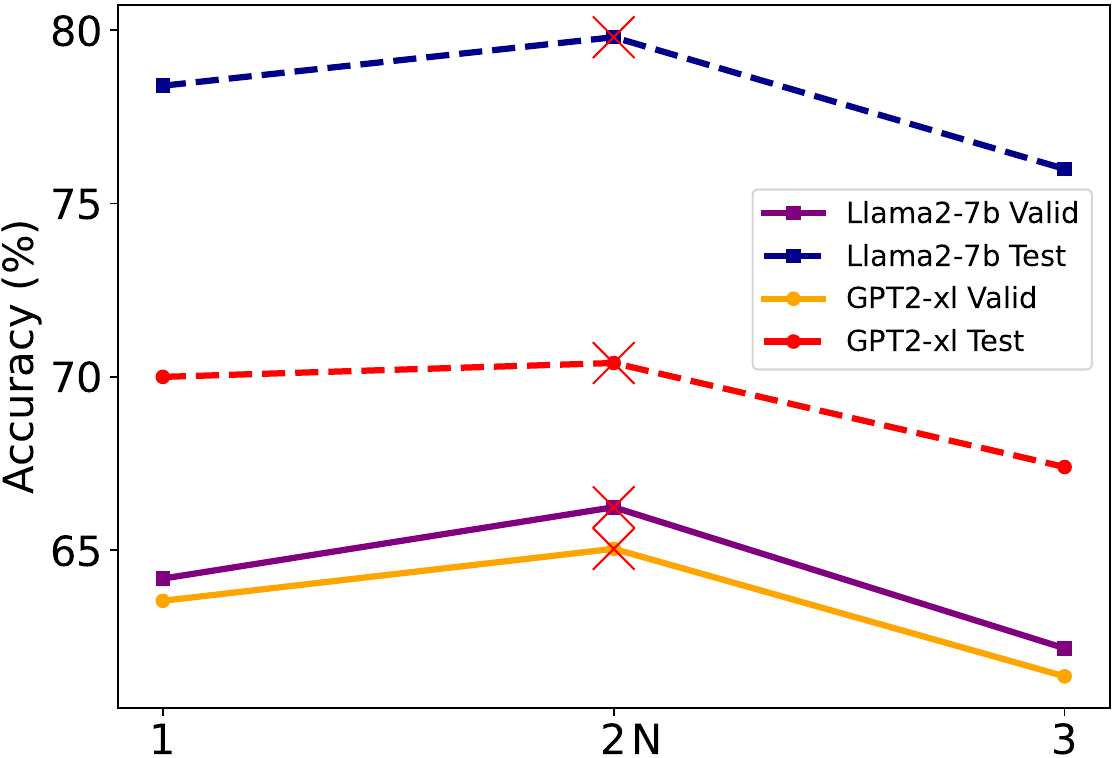}}
\subfigure[AMAN]
{\label{aman}
\includegraphics[height=2.6cm, width=0.32\textwidth]{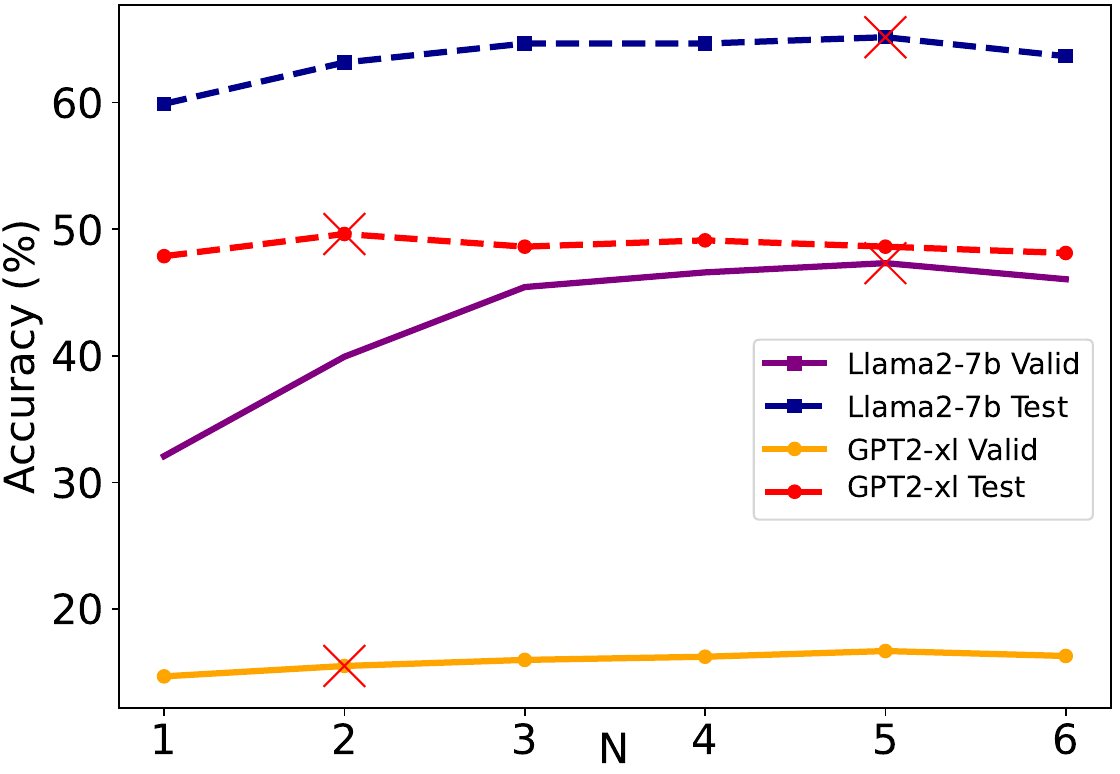}}
\vspace{-0.5cm} 
\caption{Validation and test performance under inserted word quantity (N) in sample-multiple-label pairs. The red cross marks the reported result setting. In LLaMA2-7b, N is 2 for SST2, IMDB, TREC, ISEAR, and AGNews, 4 for CR, and 5 for AMAN. In GPT2-xl, N is 7 for ISEAR and 2 for others. The remaining datasets are in Appx.\ref{appendixd}.}
\label{validationN}
\end{figure*}

\noindent\textbf{Validation-Testing Results Alignment:} Fig.\ref{validationN} displays the validation and test performance across various counts of inserted class-related words (N) for each dataset in LICL. Generally, the trends in test performance are consistent with those in validation performance. The peak test accuracy was achieved by incorporating multiple class-related words before reaching the stopping criterion, at which point the validation accuracy begins to decrease with further insertion of class-related words. This consistency supports the reliability of our stopping criterion in sequential forward insertion.

\noindent\textbf{Initial Class-Related Word Updates Improved} To boost the quality of the initial sample-label pairs, we update certain labels in two datasets based on the samples selected in LICL and DataICL before ICL experiments\footnote{Updating initial class-related words ensures they have a high logit value relative to other words with the same label polarity to provide effective class-related guidance in demonstrations for all testing samples. However, Vanilla, TopK, and SelfICL use varying demonstration samples across test instances, leading to inconsistent evaluations and making updates impractical. Therefore, we use class names as initial words in the multiple insertion experiment.}. In AMAN, `other' is replaced with `neutral'. In TREC, `entity' is replaced with `animal', `description' with `definition', `human' with `persons', `location' with `state', and `number' with `numeric'. The enhanced accuracy presented in Table \ref{quality} highlights the effectiveness of class-related word updates over the direct use of predefined class names.

\begin{table}[htbp]
\centering
\resizebox{\linewidth}{!}{
\begin{tabular}{ccccccccc}
\hline
     & \multicolumn{4}{c}{\textit{LLaMA2-7b}}                           & \multicolumn{4}{c}{\textit{GPT2-xl}  }                           \\ \hline
     & \multicolumn{2}{c}{DataICL} & \multicolumn{2}{c}{LICL}  & \multicolumn{2}{c}{DataICL} & \multicolumn{2}{c}{LICL}  \\ \cline{2-9} 
     & original  & update          & original & update         & original  & update          & original & update         \\ \cline{2-9} 
AMAN & 40.85     & \textbf{53.88}  & 57.00    & \textbf{59.90} & 31.08     & \textbf{35.34}  & 41.35    & \textbf{47.87} \\
TREC & 70.20     & \textbf{71.80}  & 70.40    & \textbf{78.40} & 37.40     & \textbf{57.20}  & 54.20    & \textbf{70.00} \\ \hline
\end{tabular}
}
\caption{Enhancement in 1-shot ICL accuracy (\%) through label evaluation and updates.}
\label{quality}
\end{table}

\subsection{Effectiveness of LICL in 5-shot ICL \label{sec:5shot}} To explore the capability of multiple class-related words further, we assess their effectiveness in a 5-shot setting using our method and SelfICL for one binary classification task (SST2) and one multi-class classification task (AMAN). For both methods, the 5-shot sample-label pairs are selected based on the top-5 scoring samples in the training set for each label. The validation set is created by removing all selected samples from the training set. The sample-label pairs are ordered based on the highest score achieved by each label among the selected samples.

\begin{table}[htbp]
\centering
\resizebox{\linewidth}{!}{
\begin{tabular}{rclccclcc}
\hline
\multicolumn{1}{c}{}                                      & SST2           & TREC  & AMAN           & Avg.     & SST2           & TREC  & AMAN           & Avg.     \\ \hline
\multicolumn{1}{l}{\textit{}}                             & \multicolumn{4}{c}{\textit{Llama2-7b}}                   & \multicolumn{4}{c}{\textit{GPT2-xl}}                     \\
\multicolumn{1}{l}{\textbf{SelfICL}}                      & 93.63          & 73.20 & 50.13          & 72.32          & 73.04          & 59.00 & 34.56          & 55.53          \\
+MLabels\_CN                                              & 94.89          & 74.00 & 55.89          & 74.93          & 86.60          & 69.20 & 40.40          & 65.38          \\
+MLabels\_LW                                           & 94.89          & 74.00 & 56.14          & 75.01          & 86.60          & 69.20 & 40.60          & 65.47          \\ \hline
\multicolumn{1}{l}{\cellcolor[HTML]{ECF4FF}\textbf{LICL}} & 94.56          & 76.20 & 56.14          & 75.35          & 81.16          & 72.00 & 34.84          & 62.67          \\
\cellcolor[HTML]{ECF4FF}+MLabels\_CN                      & 95.39          & 78.60 & 58.40          & 76.90          & 89.18          & \textbf{74.40} & 51.80          & \textbf{71.79} \\
\cellcolor[HTML]{ECF4FF}+MLabels\_LW                      & \textbf{95.39} & \textbf{78.60} & \textbf{58.40} & \textbf{76.90} & \textbf{89.18} & 73.00 & \textbf{51.80} & 71.33          \\ \hline
\end{tabular}
}
\caption{Accuracy (\%) with multiple class-related word insertion in 5-shot ICL: The scoring method and test sets are consistent with those in Table \ref{LICLoverperformance}.}
\label{5shot}
\end{table}

As shown in Table \ref{5shot}, the insertion of multiple class-related words is effective in 5-shot settings, yielding an average accuracy improvement of 2.69\% and 9.93\% in SelfICL, and 1.08\% and 9.13\% in LICL under LLaMA2-7b and GPT2-xl. LICL outperforms SelfICL across all results. Inserted words for both baseline and our models are in Appx.\ref{appendixd}, and enhancements for updating initial class-related words under 5-shot settings are in Appx.\ref{5-shot-initial}.

\subsection{Ablation Studies}

\subsubsection{Effectiveness of LICL Ordering\label{sec:permutation}}
In LICL, we use the logit separability scores of samples as a decision feature to order sample-label pairs. To assess the effectiveness of this ordering strategy, we compare 1-shot ICL classification performance using LICL’s initial sample-label pairs against 30 (LLaMA2-7b) and 50 (GPT2-xl) random permutations (excluding LICL’s order) in multi-class tasks\footnote{We randomize the order of samples selected by LICL to create various permutations and then randomly select 30 or 50 different permutations from the generated pool. Each permutation is assessed individually in a 1-shot ICL classification.}, and a flipped order in binary tasks.  

\begin{table}[htbp]
\centering
\resizebox{\linewidth}{!}{
\begin{tabular}{lcccccccc}
\hline
                     & SST2           & CR             & IMDB           & TREC           & AMAN           & ISEAR          & AGNews         & Avg.     \\ \hline
\textit{LLaMA2-7b}   &                &                &                &                &                &                &                &                \\
\textbf{LICL}        & \textbf{95.39} & \textbf{94.41} & \textbf{95.40} &\textbf{78.40}         & 59.90          & 72.49          & 84.11          & \textbf{82.87} \\
\textbf{Permutation} & 92.97          & 93.35          & 94.40          &78.20 & \textbf{60.15} & 72.49          & \textbf{85.82} & 82.48          \\ \hline
\textit{GPT2-xl}     &                &                &                &                &                &                &                &                \\
\textbf{LICL}        & \textbf{85.17} & \textbf{64.89} & 71.50          & \textbf{70.00} & \textbf{47.87} & 48.64          & \textbf{78.92} & \textbf{66.71} \\
\textbf{Permutation} & 53.87          & 63.83          & \textbf{82.20} & 63.20          & 39.10          & \textbf{52.03} & 73.16          & 61.06          \\ \hline
\end{tabular}
}
\caption{Accuracy (\%) comparison of our ordering method with random orders in multi-class datasets and the flipped order in binary datasets. `Permutation' presents the best result for each dataset.}
\label{permutation}
\end{table}
\vspace{-2mm}

As shown in Table \ref{permutation}, LICL often outperforms or matches the best results among compared permutations. It achieves an average accuracy improvement of 0.39\% in LLaMA2-7b and 5.65\% in GPT2-xl. Although GPT2-xl shows a performance drop on the IMDB dataset, the binary nature of this task can mitigate this by allowing all possible orders to be tested exhaustively. LICL excels in multi-class tasks, matching or exceeding top permutation results. This underscores LICL's ordering strategy's efficiency, especially in multi-class tasks with thousands of possible orderings ($7!$). While enumerating all these orderings could take months, LICL achieves comparable performance within hours or minutes, demonstrating its practical advantage.


\subsubsection{Effectiveness of Label Organization\label{sec:insertaba}}
In Table \ref{LICLoverperformance}, inserting multiple class-related words significantly improves ICL performance across all models. To further evaluate the effectiveness of our label organization method, we compare it with a random selection and ordering of multiple class-related words. In Table \ref{multipleaba}, the lower accuracy observed in the `+MLabels-Random' highlights the advantage of our systematic approach. These results emphasize that thoughtful organization of class-related words is crucial for boosting ICL performance. 

\begin{table}[htbp]
\centering
\resizebox{\linewidth}{!}{
\begin{tabular}{rccccccc}
\hline
\multicolumn{1}{l}{LICL LLaMA2-7b} & SST2           & CR             & IMDB           & TREC           & AMAN           & ISEAR          & AGNews         \\ \hline
+MLabels-Random                   & 94.84          & 93.62          & 94.90          & 79.40          & 57.64          & 70.56          & 80.49          \\
+MLabels\_LW                       & \textbf{95.97} & \textbf{95.15} & \textbf{95.60} & \textbf{80.60} & \textbf{69.40} & \textbf{73.09} & \textbf{86.58} \\ \hline
\end{tabular}
}
\caption{Accuracy (\%) comparison of our multiple class-related words organization method with random multiple words organization in $P_r$ (same N as in Table \ref{LICLoverperformance}).}
\label{multipleaba}
\end{table}
\vspace{-3mm}

\subsection{Scalability to Larger LLMs}\label{sec:lagerllm}
Table \ref{llama3-llama12} presents the accuracy results of LICL implemented on LLaMA3-8b and LLaMA2-13b, validating the effectiveness and applicability of our method on larger language models. The insertion of multiple class-related words improved accuracy, with LICL achieving a 3.42\% increase on LLaMA3-8b and 2.32\% on LLaMA2-13b. Notably, for the IMDB dataset on LLaMA2-13b, despite significant sample information loss compared to other methods, our approach maintained strong performance. Details of the evaluation of initial class-related word updates can be found in Appx.\ref{l-llm}.

\begin{table}[htbp]
\centering
\resizebox{\linewidth}{!}{
\begin{tabular}{lcccccccl}
\hline
\multicolumn{1}{c}{} & SST2           & CR             & IMDB            & TREC           & AMAN           & ISEAR          & AGNews         & Avgs.          \\ \hline
\textit{LLaMA3-8b}   &                &                &                 &                &                &                &                &                \\
LICL                 & 95.66          & 94.29          & 95.50           & 79.00          & 61.15          & 70.83          & 84.67          & 83.01          \\
+MLabels\_CN    & 95.77          & 95.82          & 95.90           & 80.40          & 65.89          & 73.89          & 87.72          & 85.06          \\
+MLabels\_LW    & \textbf{95.77} & \textbf{95.82} & \textbf{95.90}  & \textbf{86.00} & \textbf{69.57} & \textbf{74.29} & \textbf{87.72} & \textbf{86.44} \\ \hline
\textit{LLaMA2-13b}  &                &                &                 &                &                &                &                &                \\
LICL                 & 96.38          & 94.62          & 95.70*          & 82.20          & 60.15          & 73.16          & 85.12          & 83.90          \\
+MLabels\_CN    & 96.87          & 95.88          & 95.90*          & 84.00          & 67.64          & 74.37          & 87.79          & 86.06          \\
+MLabels\_LW    & \textbf{96.87} & \textbf{95.88} & \textbf{95.90*} & \textbf{84.40} & \textbf{67.89} & \textbf{74.82} & \textbf{87.79} & \textbf{86.22} \\ \hline
\end{tabular}}
\caption{Accuracy performance (\%) of LICL on LLaMA2-13b and LLaMA3-8b. *: For the IMDB dataset on LLaMA2-13b, our GPU could only accommodate 20\% of each demonstration sample length.}
\label{llama3-llama12}
\end{table}
\vspace{-3mm}

\subsection{General Applicability of Inserted Words} \label{sec: cross}

To address potential concerns about overfitting when selecting multiple class-related words from a single dataset, we implement cross-dataset evaluations. We evaluate the efficacy of using two class-related words (N in most binary tasks in our study) selected from the SST2 dataset across the CR and IMDB datasets, as detailed in Table \ref{crossvalidation}.

\begin{table}[htbp]
\centering
\resizebox{\linewidth}{!}{
\begin{tabular}{lcccc}
\hline
\multicolumn{1}{c}{}                                  & CR            & IMDB          & CR           & IMDB         \\ \hline
                                                      & \multicolumn{2}{c}{\textit{LLaMA2-7b}} & \multicolumn{2}{c}{\textit{GPT2-xl}} \\ \cline{2-5} 
LICL                                                  & 94.41         & 95.40         & 64.89        & 71.50        \\
\multicolumn{1}{r}{+MLabels (from same dataset)} & 95.15         & 95.60         & 69.95        & 73.40        \\
\multicolumn{1}{r}{+MLabels (from SST2)}         & 95.15         & 95.40         & 65.04        & 72.86        \\ \hline
\end{tabular}}
\caption{Cross-dataset accuracy (\%) evaluation on inserted multiple class-related words.}
\label{crossvalidation}
\end{table}

In LLaMA2-7b, performance improvements on the CR may be attributed to similar sub-token/token found in both SST2 and CR (e.g., " \textbf{un}healthy", " \textbf{un}fair", " \textbf{good}"). No performance decrease is observed in IMDB. In GPT2-xl, despite initial differences in the positive sentiment words across datasets, the class-related words from SST2 still enhanced performance in both CR and IMDB. These results indicate that the selected class-related words exhibit a level of generality, enhancing ICL performance across various datasets with similar labels.

\subsection{Further Analyses}\label{sec:futherana}
More in-depth analyses are in Appx.\ref{indepthanalysis}, including pool filtering effectiveness and its ablation study (\ref{sec:filtering}), label bias evaluation (\ref{sec: case1}), the impact of class-related words on predictions whether or not they appear in demonstrations (\ref{sec:demonstrationprediction}), the effects of label-balanced demonstration in LICL (\ref{sec:labelbalance}), sample-label logit separability visualizations (\ref{sec:visualization}), and inference cost (\ref{sec: case4}).

\section{Conclusion}
In this paper, we introduce logit separability, a criterion for measuring the instruction clarity of samples and labels based on their logit value distributions, assessing whether a sample is easily predictable by its true label and whether a label performs well across multiple samples of the same class. To improve ICL with clearer class-related guidance, we propose LICL, a logit separability-based method for selecting and ordering demonstrated samples and labels. Besides, to enrich ICL with broader label information, we present a novel label demonstration method by forming sample-label pairs through the insertion of multiple class-related words, guided by sequential forward search on the logits from selected samples and validation performance. Experimental results confirm that this combined approach significantly enhances ICL classification performance, yielding superior outcomes.

\section{Limitations}
In this paper, we enhance in-context learning performance by incorporating additional class-related words. Although related class-related words for various tasks have been extracted and collected, new datasets may still lack appropriate class-related words. However, powerful search tools such as WordNet \cite{pedersen2004wordnet}, ConceptNet \cite{speer2017conceptnet}, and open-source vocabularies can mitigate this issue. Our method designs a filtering approach that refines the quality of class-related words based on these search results.

\section*{Acknowledgements}
The research was conducted at the Future Resilient Systems at the Singapore-ETH Centre, and is supported by the National Research Foundation Singapore under its Campus for Research Excellence and Technological Enterprise programme.

\bibliography{acl_latex}

\appendix
\section{Exploring Logit Separability and Class-Related Word Effectiveness}\label{appendixa}

\subsection{Logit Separability Over Class-Related Words in Zero-Shot Learning}\label{appendixa2}
\begin{figure}[htbp]
\centering  
\subfigure[`pessimistic' vs `negative']{
\label{pessimistic}
\includegraphics[width=0.43\textwidth]{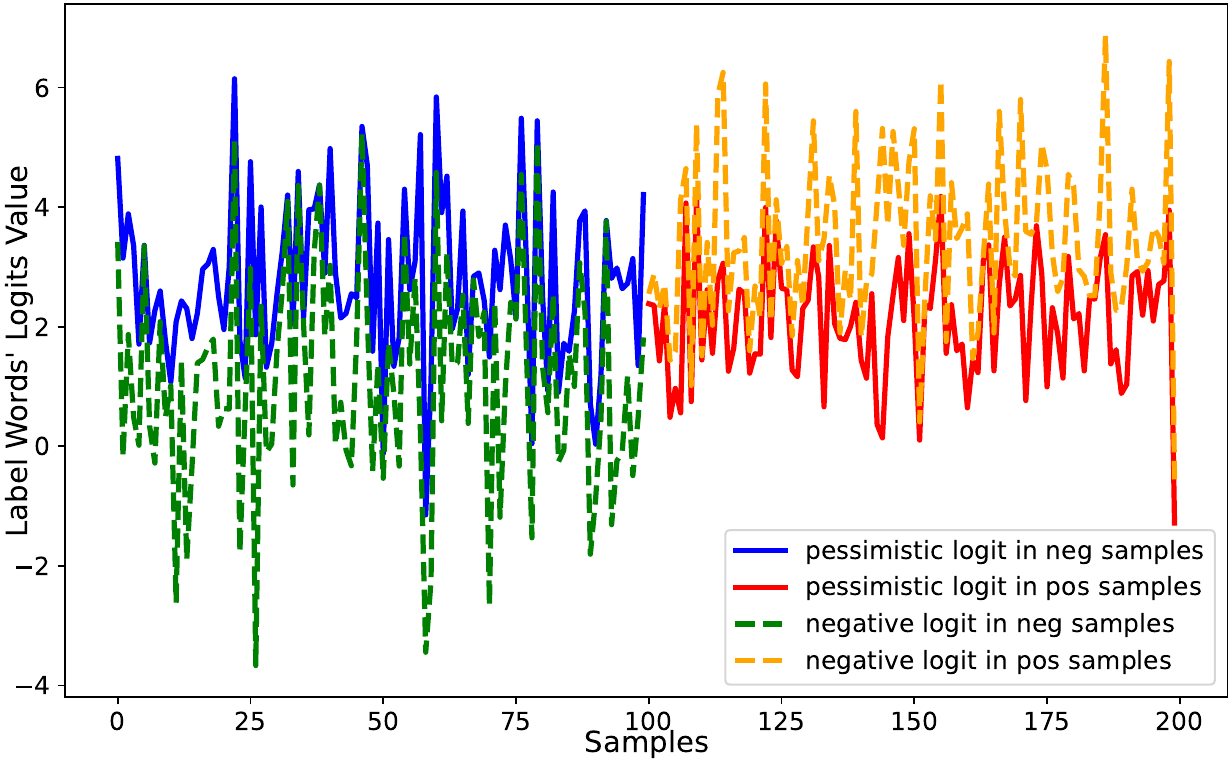}}
\subfigure[`good' vs `positive']{
\label{good}
\includegraphics[width=0.43\textwidth]{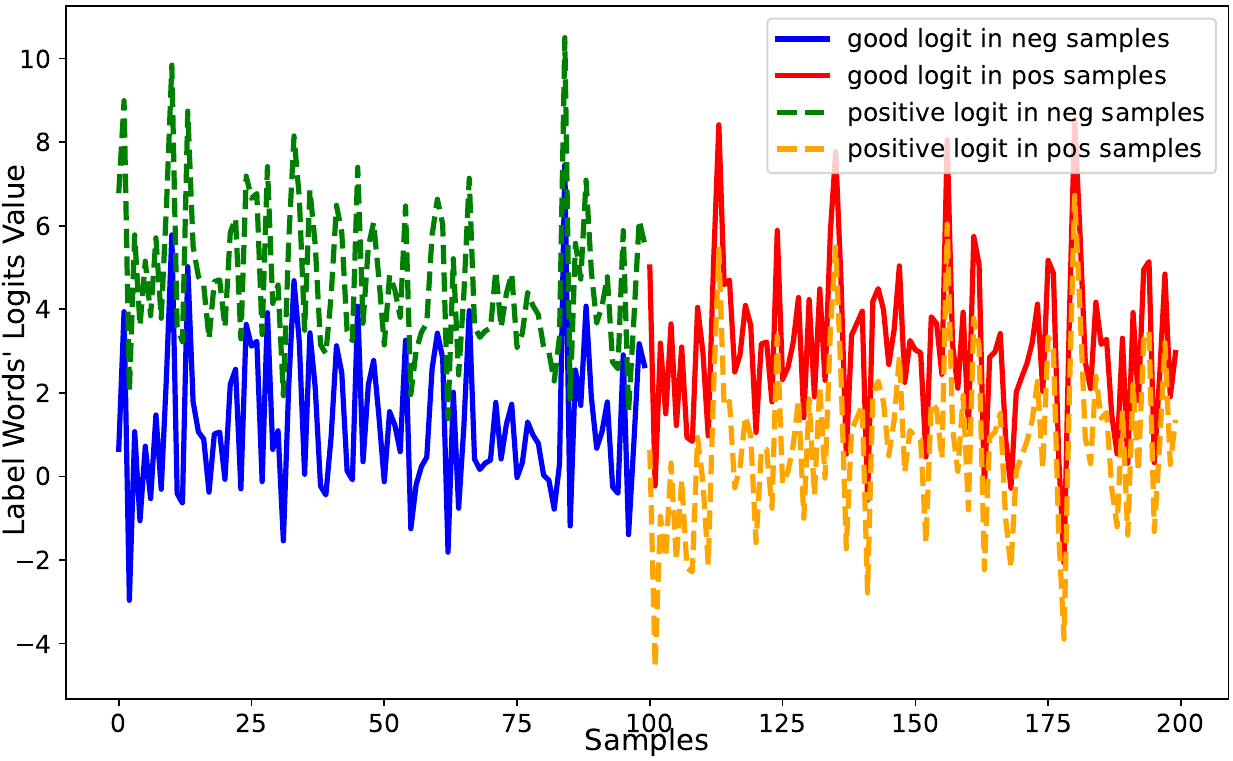}}
\subfigure[`happy' vs `positive']{
\label{happy}
\includegraphics[width=0.43\textwidth]{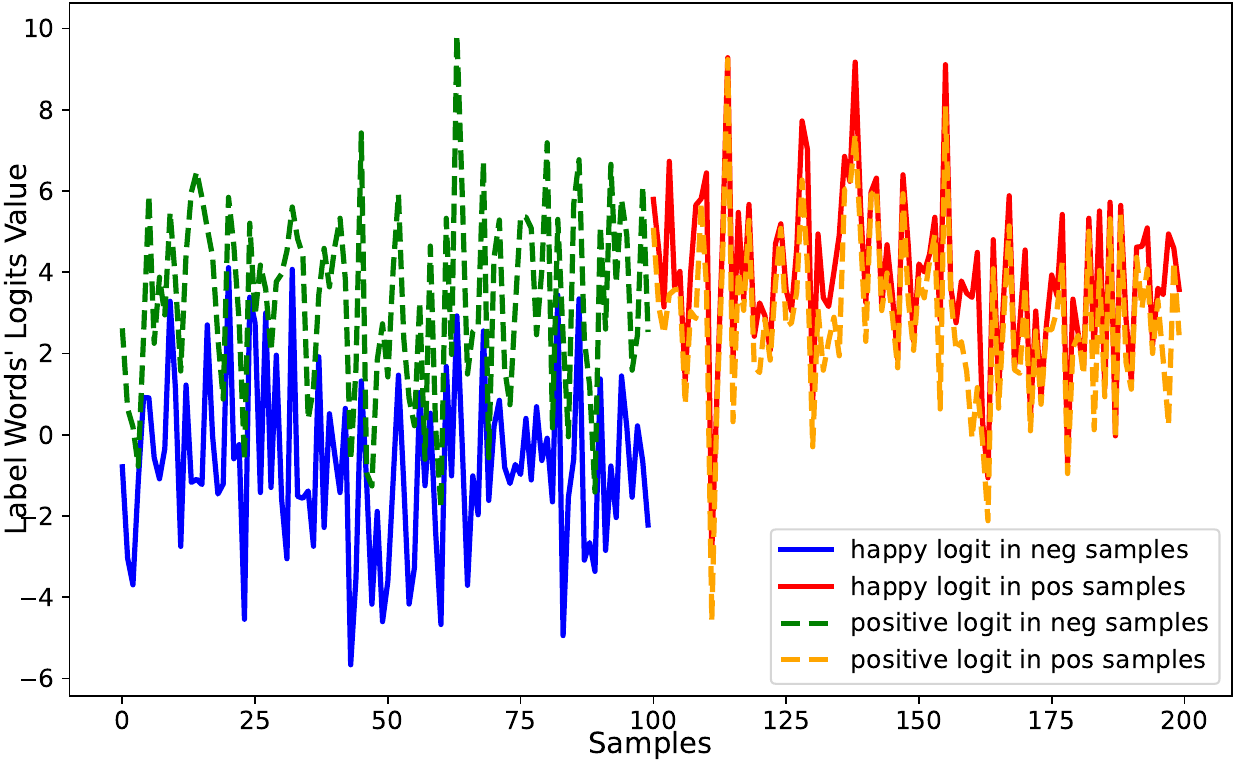}}
\caption{class-related words logit separability over samples.}
\label{otherdistribution}
\end{figure}

This study evaluates the logit value separability for negative and positive samples of class-related words `bad' and `pessimistic' compared to the class name `negative', and for class-related words `good' and `happy' compared to the class name `positive' in SST-2 under zero-shot learning. Except for the word bad', shown in Fig.\ref{distribution}, the logit distribution figures for the remaining words are listed in Fig \ref{otherdistribution}.

\begin{table*}[htbp]
\centering
\resizebox{\linewidth}{!}{\begin{tabular}{cccccc}
\hline
Label Set & SST2                                                                      & CR                                                                       & TREC                                                                                                                                      & AMAN                                                                                                                                              & ISEAR                                                                                                                                                                   \\ \hline
1         & \begin{tabular}[c]{@{}c@{}}0: `0',\\ 1: `1'\end{tabular}                  & \begin{tabular}[c]{@{}c@{}}0: `0',\\ 1: `1'\end{tabular}                 & \begin{tabular}[c]{@{}c@{}}0: `0', 1: `1', 2: `2',\\  3: `3', 4: `4', 5: `5'\end{tabular}                                                         & \begin{tabular}[c]{@{}c@{}}0: `0', 1: `1', 2: `2',\\  3: `3', 4: `4', 5: `5', 6: `6'\end{tabular}                                                 & \begin{tabular}[c]{@{}c@{}}0: `0', 1: `1', 2: `2',\\  3: `3', 4: `4', 5: `5', 6: `6'\end{tabular}                                                                       \\
2         & \begin{tabular}[c]{@{}c@{}}0: ` negative', \\ 1: ` positive'\end{tabular} & \begin{tabular}[c]{@{}c@{}}0: ` negative',\\ 1: ` positive'\end{tabular} & \begin{tabular}[c]{@{}c@{}}0: ` abbreviation', 1: ` entity', 2: ` description',\\  3: ` human', 4: ` location',5: ` number'\end{tabular} & \begin{tabular}[c]{@{}c@{}}0: ` fear', 1: ` sadness', 2: ` disgust', 3: ` anger',\\  4: ` joy', 5: ` surprise', 6: ` others'\end{tabular}         & \begin{tabular}[c]{@{}c@{}}0: ` fear', 1: ` sadness', 2: ` disgust', 3: ` anger',\\  4: ` joy', 5: ` guilt', 6: ` shame'\end{tabular}                                   \\
3         & \begin{tabular}[c]{@{}c@{}}0: ` bad',\\ 1: ` good'\end{tabular}           & \begin{tabular}[c]{@{}c@{}}0: ` bad',\\ 1: ` good'\end{tabular}          & \begin{tabular}[c]{@{}c@{}}0: ` abbreviation', 1: ` animal', 2: ` definition',\\  3: ` persons', 4: ` state',5: ` numeric'\end{tabular}   & \begin{tabular}[c]{@{}c@{}}0: ` worry', 1: ` sadness', 2: ` loathing', 3: ` rage',\\  4: ` happy', 5: ` stunning', 6: ` neutral'\end{tabular}     & \begin{tabular}[c]{@{}c@{}}0: ` worry', 1: ` grief', 2: ` loathing', 3: ` rage',\\  4: ` happy', 5: ` remorse', 6: ` embarrassment'\end{tabular}                        \\
4         & \begin{tabular}[c]{@{}c@{}}0: ` terrible',\\ 1: ` great'\end{tabular}     & \begin{tabular}[c]{@{}c@{}}0: ` terrible',\\ 1: ` great'\end{tabular}    & \begin{tabular}[c]{@{}c@{}}0: ` abbreviation', 1: ` food', 2: ` reason',\\  3: ` persons', 4: ` city', 5: ` count'\end{tabular}           & \begin{tabular}[c]{@{}c@{}}0: ` anxiety', 1: ` sad', 2: ` disgusting', 3: ` angry',\\  4: ` pleasure', 5: ` surprising', 6: ` noemo'\end{tabular} & \multicolumn{1}{l}{\begin{tabular}[c]{@{}l@{}}0: ` anxiety', 1: ` sad', 2: ` disgusting', 3: ` angry',\\  4: ` pleasure', 5: ` regret', 6: ` humiliation'\end{tabular}} \\ \hline
\end{tabular}}
\caption{The label information of each dataset.}
\label{labelinfo}
\end{table*}

In Fig.\ref{distribution} and Fig.\ref{otherdistribution}, the first 100 samples are negative, while samples 101-200 are positive samples. Compared to the class names, the logit of negative class-related words across negative samples is higher than those for `negative', while the logit for the same class-related words across positive samples is lower than `negative'. Similarly, the logit for positive class-related words is higher across positive samples and lower across negative samples than `positive', indicating better logit separability for these class-related words compared to their respective class names. This demonstrates superior logit separability for certain class-related words compared to class names. Since logit values are crucial for class prediction, this enhanced separability can significantly improve classification performance.

\subsection{Label Effectiveness in ICL}
\subsubsection{Single Class-Related Word as Demonstration Label}\label{appendixa1}
\begin{figure}[hpbt]
\centering
\includegraphics[height=7cm, width=0.4\textwidth]{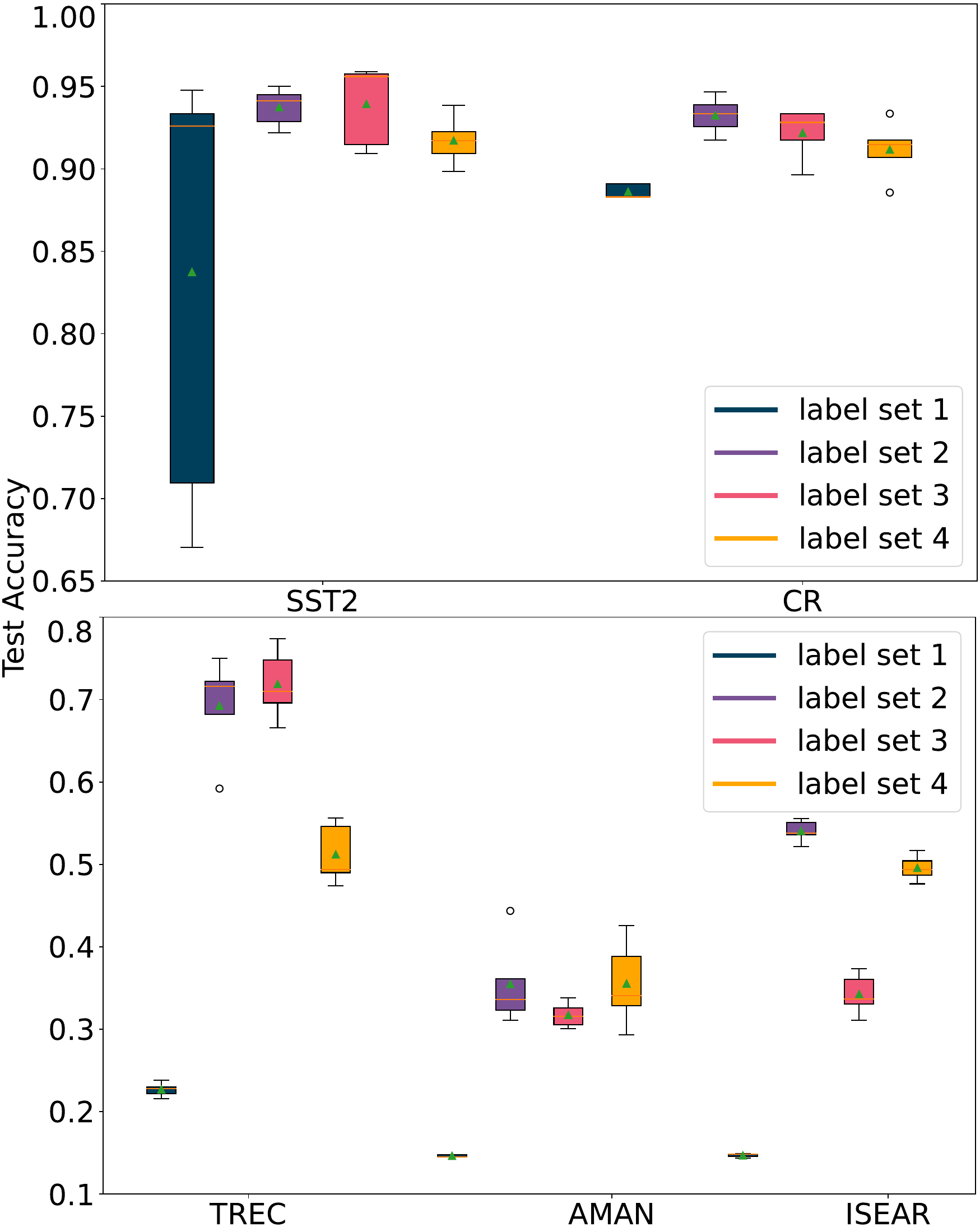}
\caption{Label effectiveness in ICL (GPT2-xl)}
\label{effectivenessllama2}
\end{figure}

\begin{figure}[hpbt]
\centering
\includegraphics[height=7cm, width=0.4\textwidth]{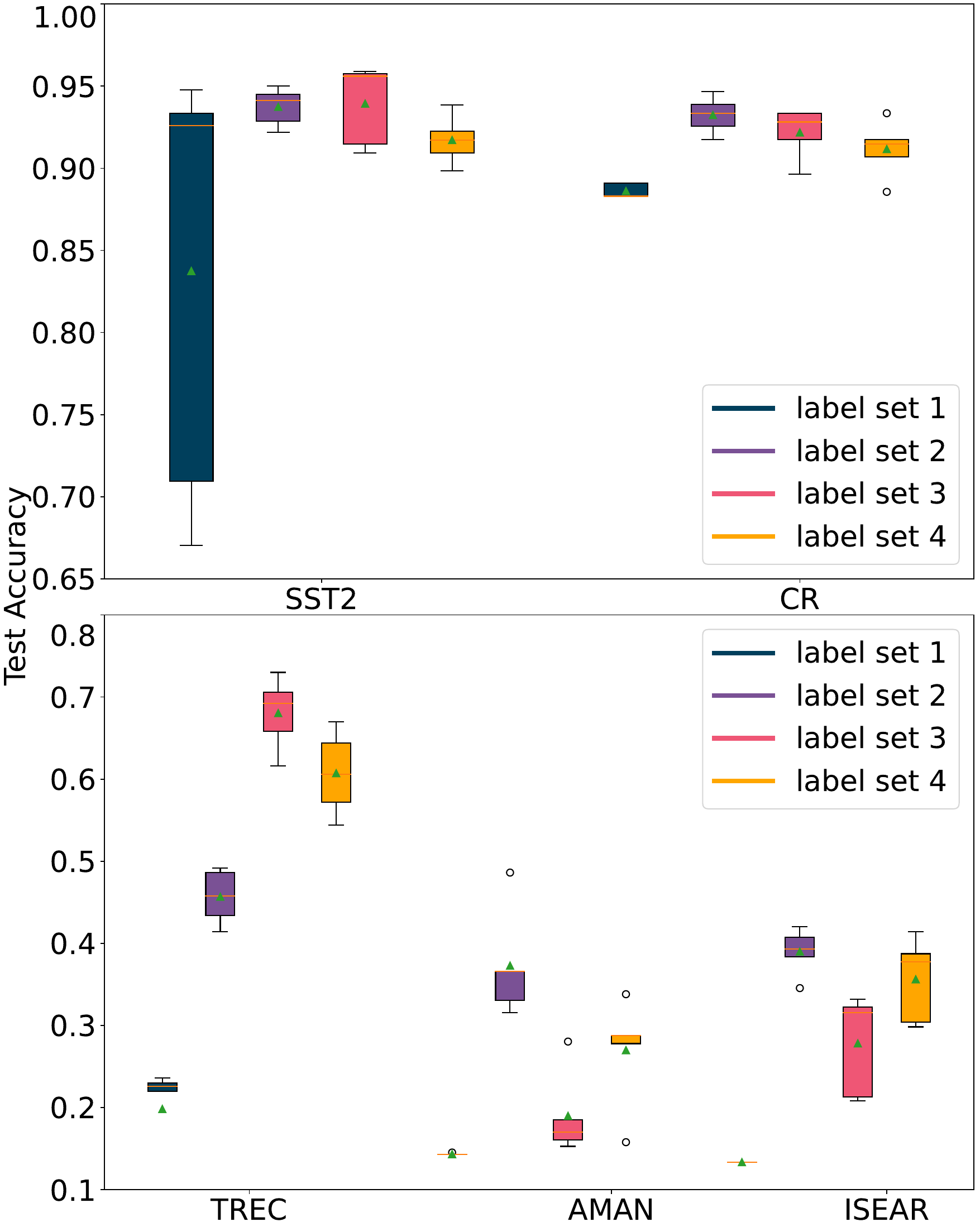}
\caption{Label effectiveness in ICL (GPT2-xl)}
\label{effectivenessgpt2}
\end{figure}

This study establishes four distinct label sets for each dataset, utilizing identical samples to form the sample-label pairs in 1-shot ICL, as detailed in Table \ref{labelinfo}. The reported results represent the average accuracy obtained from five repeated experiments, conducted with seeds 42, 43, 44, 45, and 46 in sample selection during the 1-shot demonstrations. As shown in Fig.\ref{effectivenessllama2}, in LLaMA2-7b, the maximum 49.20\% accuracy difference (TREC), and the highest 11.08\% (SST2) standard deviation are observed. We also evaluate the label effectiveness in GPT2-xl under the same experimental conditions, with results shown in Fig.\ref{effectivenessgpt2}. The results are similar to those in LLaMA2-7b, with a maximum accuracy difference of 48.20\% (TREC) and a maximum standard deviation of 10.52\% (SST2). These findings suggest that certain class-related words with stronger logit separability, as exampled in Figs.\ref{distribution} and \ref{otherdistribution}, can serve as more effective demonstration labels, contributing to greater accuracy and robustness in ICL than predefined class names.


\subsubsection{Multiple Class-Related Words as Demonstration Label}\label{appendixa3}
\begin{figure}[hpbt]
\centering
\includegraphics[scale=0.4]{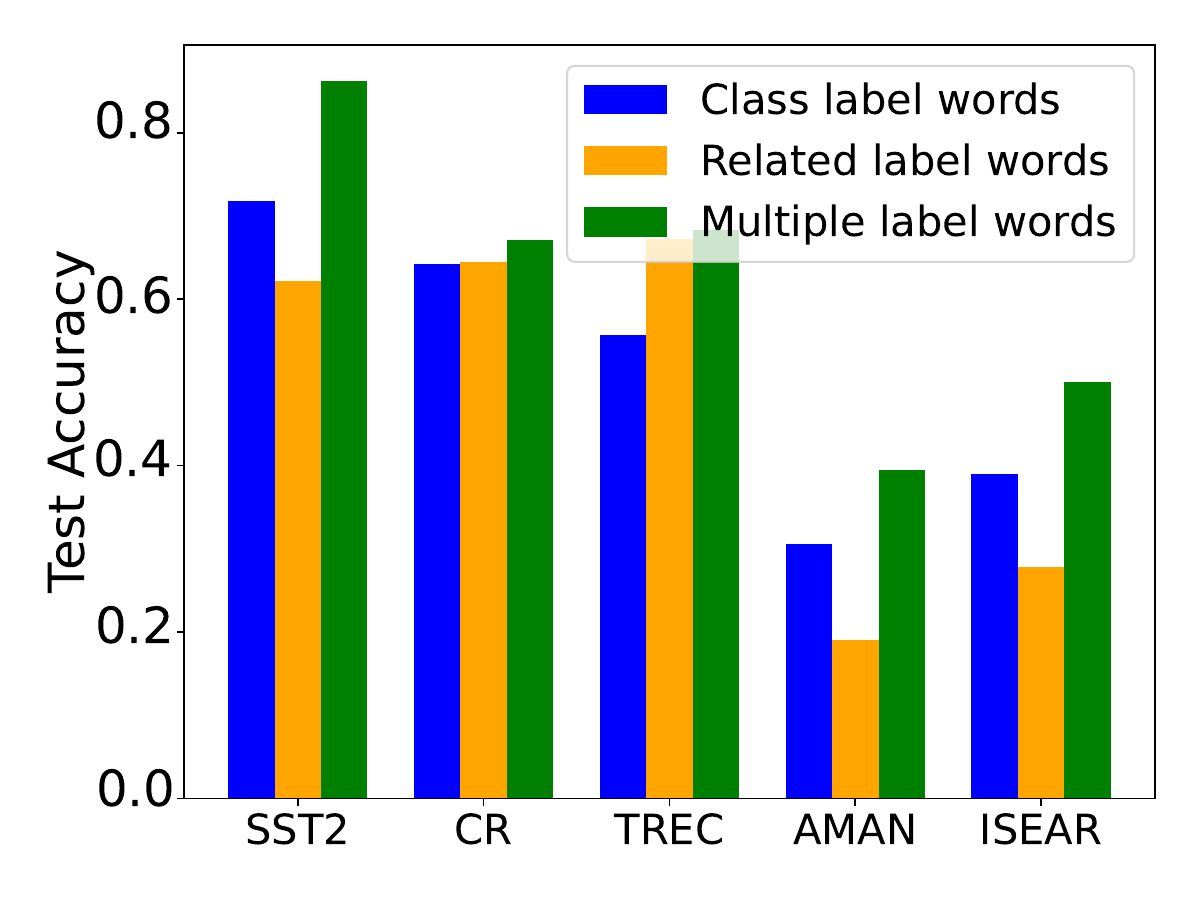}
\caption{Multiple class-related words effectiveness in ICL (GPT2-xl)}
\label{Meffectiveness}
\end{figure}

This study evaluates the performance of demonstrations using different numbers of class-related words. The multiple class-related words combine the class name with related class-related words, connected by spaces, such as "negative bad" and "positive good" in SST2 and CR. We also assess the label effectiveness in GPT2-xl under the same experimental conditions, with the results shown in Fig.\ref{Meffectiveness}. These findings indicate the potential of using multiple class-related words in demonstrations to enhance ICL.

\section{Class-Related Word Pool Filtering}\label{sec: labelwordfilter}
The goal of class-related word pool filtering is to enhance the logit separability of the words within the pool, guided by logit feedback from LLM across the dataset. Ideally, each word should exhibit a clearly dominant logit value in samples with the corresponding label, compared to those with different labels, to prevent the introduction of incorrect class-related features in the ICL demonstration.

To achieve this, we first evaluate the separability of each class-related word based on its logit value across the training set $D$ in a zero-shot setting, where input samples are formatted using a template without labels. For a word $w \in P$ with label $l \in L$, we exclude it if its average logit $b_l$ for training samples labeled with $l$ is not the highest among all training samples. This step ensures that the class-related word provides accurate logit feedback in the corresponding training samples compared to those with different label expressions. To further ensure the separability of the feedback, avoiding situations where, for example, a class-related word has an average logit of 7.1 in correct training samples but a close value of 7.0 in semantically unrelated samples, we apply point-biserial testing to test the correlation $r$ between the word's logit values in samples with the same label versus those with different labels as defined in Eq.\ref{eq:rpb}:
\begin{equation}
r=\frac{V_0-V_1}{\sigma }\times \sqrt{\frac{n_0\times n_1}{(n_0+n_1)^{2}}}
	\label{eq:rpb}
	\end{equation} 	
where $V_0$ denotes the mean logit values of the class-related word for training samples with the same label, $V_1$ denotes the mean logit values for samples with different labels, $\sigma$ represents the standard deviation of the logit values, and $n_0$ and $n_1$ correspond to the number of samples with the same and different labels, respectively.

For a given class-related word, a positive correlation $r>0$ indicates that its average feedback across training samples within its own class significantly differs from that of samples in other classes, warranting its retention. Otherwise, the word will be removed from the pool. This process ensures that the remaining words in the refined pool $P_r$ are not only logit separable but also exhibit clear class-related information distinctions.

\section{Prompt Examples}\label{appendixprompts}
The prompts used in this paper are exampled below:

\noindent \textbf{Zero-Shot Learning}:
\begin{quote}
\begin{verbatim}
Review: they 're easy to use
Sentiment:
\end{verbatim}
\end{quote}

\noindent \textbf{One-Shot Learning}:
\begin{quote}
\begin{verbatim}
Review: norton support is completely 
pathetic
Sentiment: negative 
Review: overall , i am very pleased 
with it
Sentiment: positive 
Review: they 're easy to use
Sentiment:
\end{verbatim}
\end{quote}

\noindent \textbf{One-Shot Multiple Class-Related Word Insertion Learning} (result from CR in LlaMA2-7b):
\begin{quote}
\begin{verbatim}
Review: it does not only have 
difficulty playing jpegs , it even 
has trouble ...
Sentiment: negative unhealthy unjust
Review: about the product the zen 
micro is a sleek , stylish ...
Sentiment: positive good favorable
Review: they 're easy to use
Sentiment:
\end{verbatim}
\end{quote}

\section{Statistic Information of Datasets and Templates}\label{appendixdatasets}
The statistic information of datasets and templates are listed in Table \ref{The hypotheses used in each dataset.}. Suppose the original dataset has no train/test split. In that case, a testing set is randomly selected, comprising 20\% of the entire dataset with a balanced-label distribution, while the remaining data is used for training (AMAN, ISEAR). If the dataset includes a validation set, the original training and validation sets are combined to form a complete training set for demonstration selection (SST2). For AGNews, only 4,000 training samples are selected, with 1,000 samples per label, due to memory constraints.

The class-related word pools for SST2, CR, IMDB, and AGNews are derived from \citet{hu2022knowledgeable}, while those for TREC, AMAN, and ISEAR are sourced from \citet{zhu2024edentail}.

\begin{table}[htbp]
\centering
\resizebox{\linewidth}{!}{\begin{tabular}{cccccc}
\hline
Dataset & Template                                                         & Class Name                                                                                               & \#Train & \#Validation & \#Test \\ \hline
SST2    & \begin{tabular}[c]{@{}c@{}}Review:\\ Sentiment:\end{tabular}     & positive, negative                                                                                        & 6920    & 872   & 1821   \\ \hline
CR      & \begin{tabular}[c]{@{}c@{}}Review:\\ Sentiment:\end{tabular}     & positive, negative                                                                                        & 3394    & -     & 377    \\ \hline
IMDB    & \begin{tabular}[c]{@{}c@{}}Review:\\ Sentiment:\end{tabular}     & positive, negative                                                                                        & 1000    & -     & 1000   \\ \hline
AMAN    & \begin{tabular}[c]{@{}c@{}}Review:\\ Emotion:\end{tabular}       & \begin{tabular}[c]{@{}c@{}}angry, disgust, joy, others,\\  surprise, sad, and fear\end{tabular}           & 4090    & -     & -      \\ \hline
ISEAR   & \begin{tabular}[c]{@{}c@{}}Review:\\ Emotion:\end{tabular}       & \begin{tabular}[c]{@{}c@{}}angry, disgust, joy, shame,\\  guilt, sadness, and fear\end{tabular}           & 7666    & -     & -      \\ \hline
TREC    & \begin{tabular}[c]{@{}c@{}}Question:\\ Answer Type:\end{tabular} & \begin{tabular}[c]{@{}c@{}}location, number, description,\\  entity, human, and abbreviation\end{tabular} & 5451    & -     & 490    \\ \hline
AGNews  & \begin{tabular}[c]{@{}c@{}}Article:\\ Answer:\end{tabular}       & \begin{tabular}[c]{@{}c@{}}Worlds, Business, \\ Sports, and Technology\end{tabular}                       & 120000  &   -    & 7600   \\ \hline
\end{tabular}}
\caption{The applied template and statistic information in each dataset.}
\label{The hypotheses used in each dataset.}
\end{table}

\section{Experimental Implementation  Details}\label{appendixsettings}
All experiments are implemented under Python 3.8 environment and PyTorch 2.1.0. with Cuda version 11.8, GPU NVIDIA RTX A5000.

\noindent\textbf{Baseline Model Experimental Settings}
The detailed information on the baseline models and the corresponding experimental settings for few-shot learning experiments is provided below. For all experiments conducted with LLaMA2-7b, LLaMA3-13b, and LLaMA3-8b, the model is configured to operate under a 4-bit setting.

\textbf{Vanilla LLaMA2-7b} \cite{touvron2023llama}: We use LLaMA2-7b\footnote{\url{https://huggingface.co/meta-llama/Llama-2-7b-chat-hf}\label{fn:llama}}, a 7 billion parameter language model with 4096 tokens available. Prompts exceeding the model's token limit are truncated in the few-shot settings. The demonstrations are randomly selected and ordered on each label using five random seeds: 42, 43, 44, 45, and 46. The reported results are the average ICL accuracy over five runs. 

\textbf{Vanilla GPT2-xl} \cite{radford2019language}: We use GPT2-xl\footnote{\url{https://huggingface.co/openai-community/gpt2-xl}\label{fn:gpt}}, a 1.5 billion parameter language model with 1024 tokens available. Prompts exceeding the model's token limit are truncated. The demonstration and results settings are the same as Vanilla LLaMA2-7b.

\textbf{TopK} \cite{liu2022makes}: An unsupervised method selects the nearest neighbors of the test samples as the demonstration samples using S-BERT\footnote{\url{https://huggingface.co/sentence-transformers/all-mpnet-base-v2}}. In the re-run experiment, we choose samples for each label in the order ranked by their semantic similarity to the test sample.

\textbf{SelfICL} \cite{wu2023self}: A supervised method selects demonstration samples via S-BERT and ranks them based on Minimum Description Length (MDL). In the re-run experiment, after selecting the candidates, we randomly choose 30 combinations (the default setting) containing one sample for each label for MDL ranking with a window size 10. The best results are used as the selected-and-ranked demonstrations for ICL testing.

\textbf{DataICL} \cite{chang2023data}: A supervised method trains a linear regressor to fit the LLM's output based on which sample is present and its order in the demonstration. In the re-run experiment, we select the sample with the highest score in each label as the demonstration samples, following the resulting order. The LLM used in DataICL is the same as the ICL evaluation model.

Table \ref{methodcompare} indicates whether each baseline and our proposed method incorporate functions for the selection and ordering of samples and labels, demonstrating that our one-stop method encompasses all components of demonstration organization.

\begin{table}[htbp]
\centering
\resizebox{\linewidth}{!}{
\begin{tabular}{lcccc}
\hline
           & \multicolumn{2}{c}{Sample}                            & \multicolumn{2}{c}{Label}                             \\ \cline{2-5} 
           & Selection                 & Ordering                  & Selection                 & Ordering                  \\ \hline
Vanilla    & X                         & X                         & X                       & -                         \\
TopK       & \checkmark & \checkmark & X                         &-                         \\
SelfICL    & \checkmark & \checkmark & X                         & -                        \\
DataICL    & \checkmark & X & X                         &-                         \\
LICL(ours) & \checkmark & \checkmark & \checkmark & \checkmark \\ \hline
\end{tabular}
}
\caption{Assessment of Demonstration Organization Methods in the Selection and Ordering of Samples and Labels. A `\checkmark' indicates that the method incorporates this feature, an `X' signifies its absence, and a `-' indicates that the feature is not applicable. }
\label{methodcompare}
\end{table}

\section{The Number of Inserted Labels Settings}\label{appendixd}
The number of class-related words (N) used in Table \ref{LICLoverperformance} on the baseline models are summarized in Table \ref{numberofword}. 

\begin{table}[htbp]
\centering
\resizebox{\linewidth}{!}{
\begin{tabular}{lccccccc}
\hline
                                       & SST2 & CR & IMDB & TREC & AMAN & ISEAR & AGNews \\ \hline
\multicolumn{1}{l}{\textit{LLaMA2-7b}} &      &    &      &      &      &       &        \\
\textbf{vanilla-LLaMA2-7b}             & 2    & 5  & 4    & 2    & 5    & 6     & 5      \\
\textbf{TopK}                          & 2    & 2  & 2    & 2    & 6    & 2     & 2      \\
\textbf{SelfICL}                       & 2    & 2  & 2    & 2    & 4    & 3     & 2      \\
\textbf{DataICL}                       & 2    & 2  & 2    & 2    & 3    & 6     & 6      \\
\textbf{LICL}                          & 2    & 4  & 2    & 2    & 5    & 2     & 2      \\ \hline
\multicolumn{1}{l}{\textit{GPT2-xl}}   &      &    &      &      &      &       &        \\
\textbf{vanilla-GPT2-xl}               & 3    & 2  & 2    & 4    & 3    & 6     & 2      \\
\textbf{TopK}                          & 3    & 3  & 2    & 4    & 6    & 5     & 2      \\
\textbf{SelfICL}                       & 3    & 2  & 2    & 3    & 6    & 5     & 2      \\
\textbf{DataICL}                       & 2    & 2  & 2    & 2    & 2    & 4     & 2      \\
\textbf{LICL}                          & 2    & 2  & 2    & 2    & 2    & 7     & 2      \\ \hline
\end{tabular}
}
\caption{The number of class-related words (N) inserted in the demonstration in the baseline models and LICL (ours) under each dataset in 1-shot ICL.}
\label{numberofword}
\end{table}

\begin{figure}[t]
\centering  
\subfigure[CR]{
\label{cr}
\includegraphics[height=3.1cm,width=0.4\textwidth]{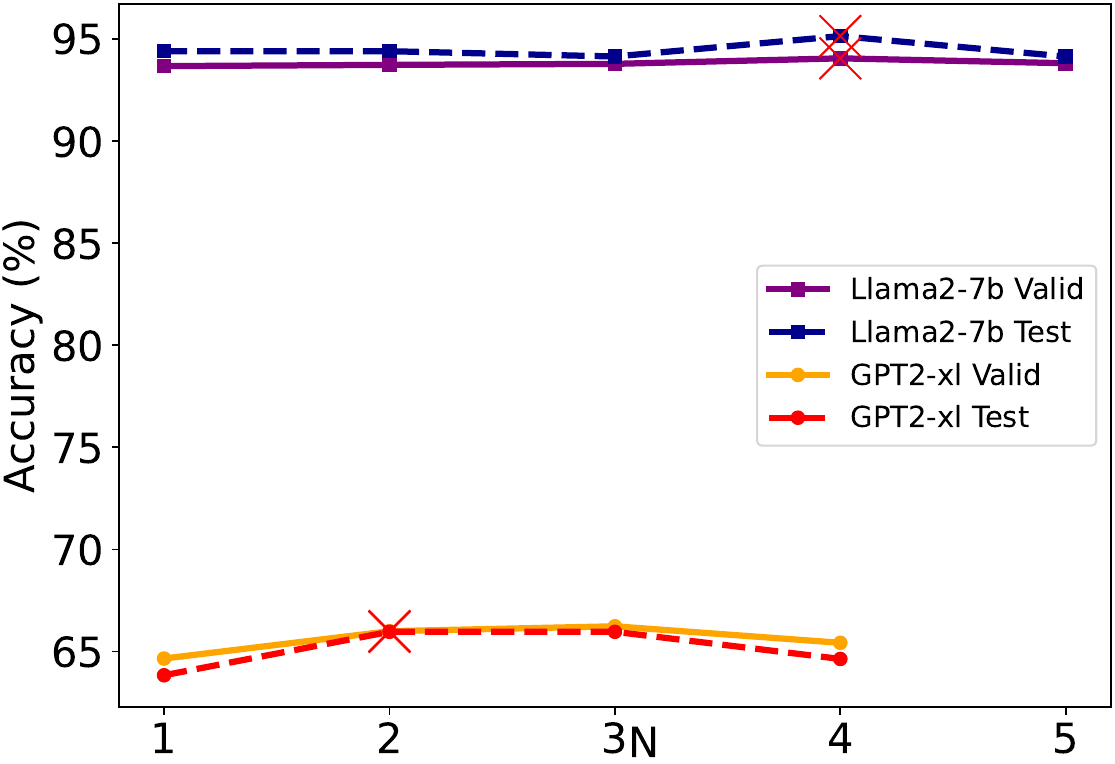}}
\subfigure[IMDB]{
\label{IMDB}
\includegraphics[height=3.1cm, width=0.4\textwidth]{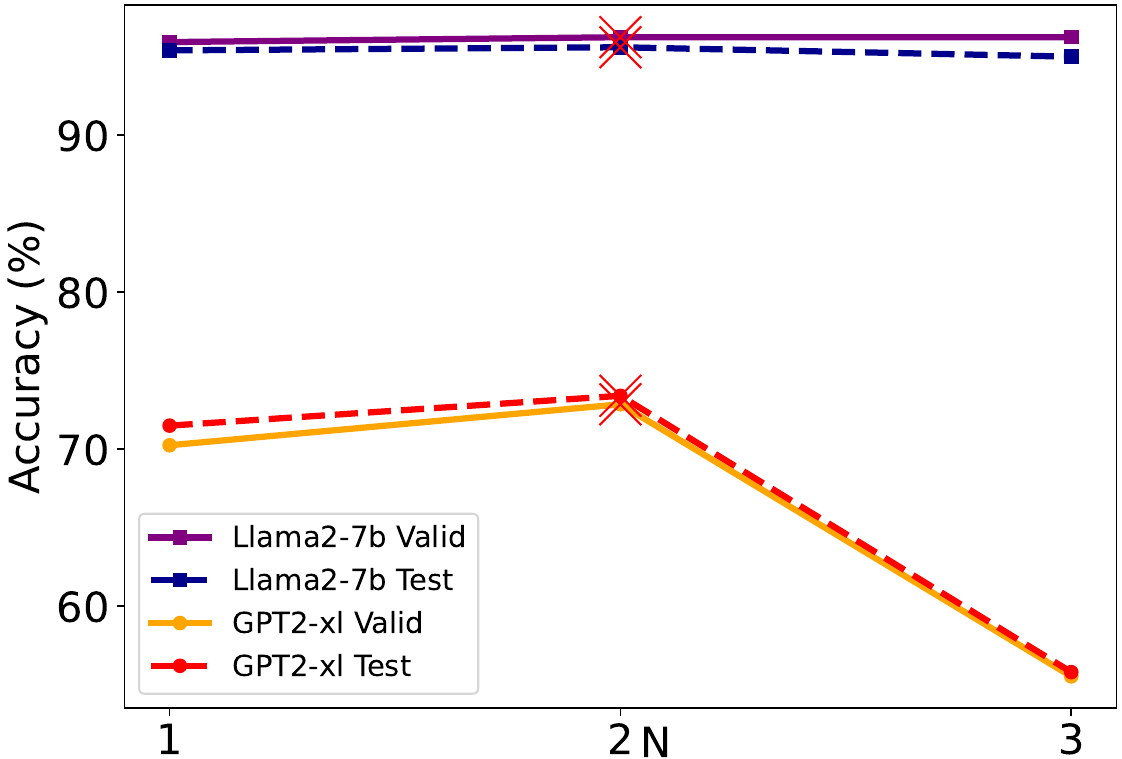}}
\subfigure[ISEAR]{
\label{ISEAR}
\includegraphics[height=3cm,width=0.4\textwidth]{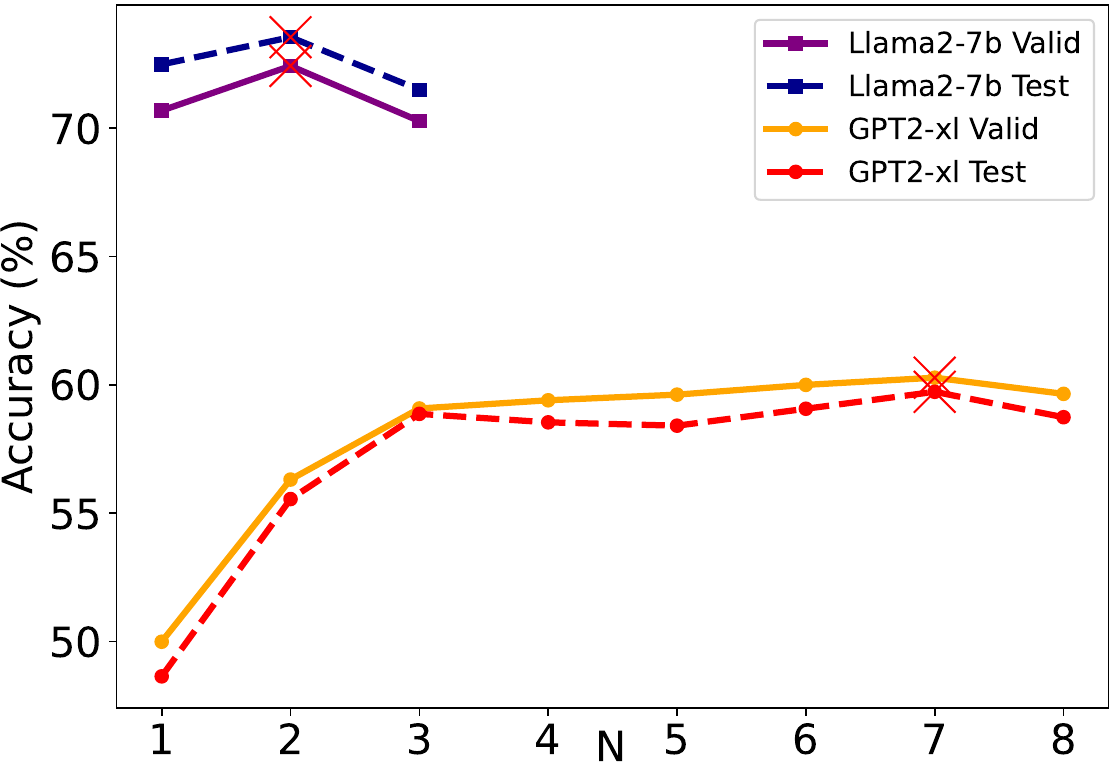}}
\subfigure[AGNews]{
\label{AGNews}
\includegraphics[height=3cm,width=0.4\textwidth]{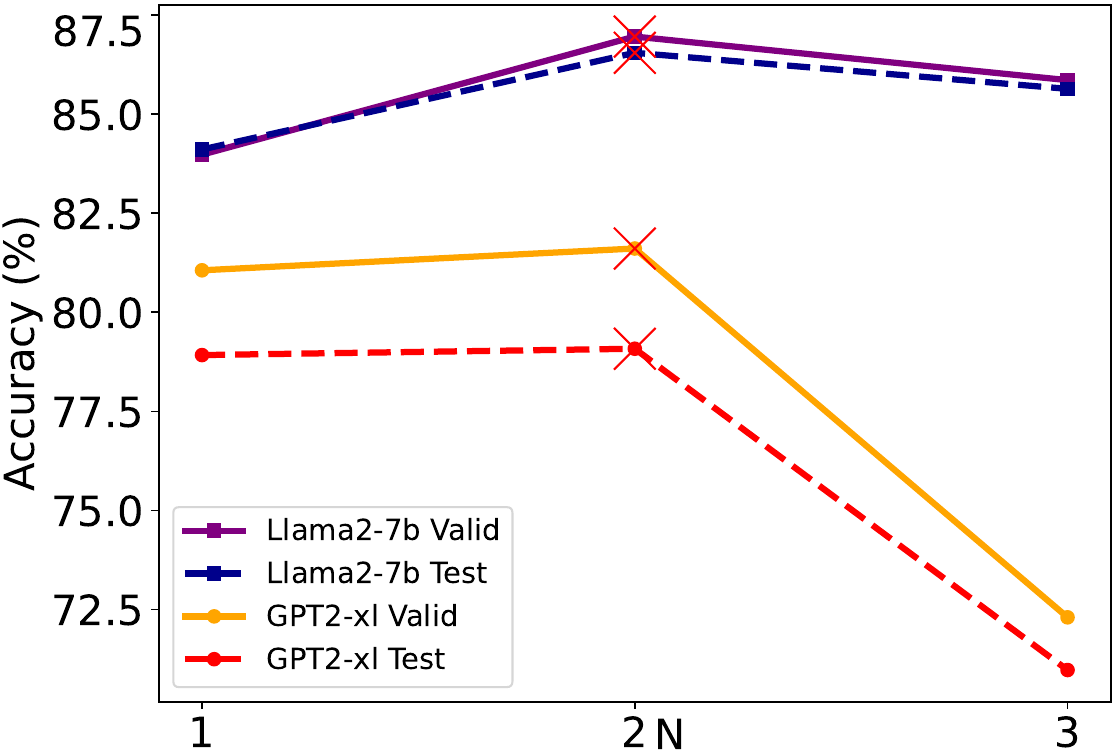}}
\caption{Validation and test performance under different class-related word quantity (N) in sample-multiple-label pairs for CR. IMDB, ISEAR and AGNews.}
\label{othervalidation}
\end{figure}

The remaining datasets' validation and test accuracy performance like Fig.\ref{validationN} is shown in Fig.\ref{othervalidation}.

The number of class-related words (N) used in Table \ref{5shot} on the baseline models are summarized in Table \ref{numberofword5shot}.
\begin{table}[htbp]
\tiny
\centering
\resizebox{\linewidth}{!}{
\begin{tabular}{lcccc}
\hline
\multicolumn{1}{c}{} & SST2               & AMAN              & SST2              & AMAN             \\ \hline
\textit{}            & \multicolumn{2}{c}{\textit{LLaMA2-7b}} & \multicolumn{2}{c}{\textit{GPT2-xl}} \\
\textbf{SelfICL}     & 2                  & 6                 & 4                 & 4                \\
\textbf{LICL}        & 2                  & 3                 & 3                 & 5                \\ \hline
\end{tabular}
}
\caption{The number of class-related words (N) inserted in the demonstration in SelfICL and LICL (ours) under SST2 and AMAN in 5-shot ICL.}
\label{numberofword5shot}
\end{table}
\normalsize

For simple classification tasks (binary tasks) such as SST2, CR, and IMDB, inserting around 2 to 3 class-related words yields good performance. In contrast, for fine-grained tasks (multi-class tasks) such as AMAN, ISEAR, and AGNews, inserting more class-related words is necessary to achieve better performance. Additionally, the larger language model (LLaMA2-7b) can effectively handle more label information compared to the smaller language model (GPT2-xl).

\section{Enhancement via Initial Class-Related Word Updates}
\subsection{5-Shot ICL}\label{5-shot-initial}

Table \ref{5shotinitial} presents the effectiveness of initial class-related word updates in 5-shot experiments on the AMAN and TREC datasets, using the same word update information as in the 1-shot setting, as described in Sec.\ref{sec:mainresult}.

\begin{table}[htbp]
\centering
\resizebox{\linewidth}{!}{
\begin{tabular}{lcclcc}
\hline
\multicolumn{1}{c}{LICL}   & AMAN               & TREC              &  & AMAN              & TREC             \\ \hline
\textit{}                  & \multicolumn{2}{c}{\textit{\textit{LLaMA2-7b}}} &  & \multicolumn{2}{c}{\textit{\textit{GPT2-xl}}} \\
Original                   & 42.35              & 70.00                 &  & 30.83             & 64.60            \\
\multicolumn{1}{r}{Update} & 56.14              & 76.20             &  & 34.84             & 72.00            \\ \hline
\end{tabular}}
\caption{Enhancement in 5-shot ICL accuracy (\%) through label evaluation and updates.}
\label{5shotinitial}
\end{table}

\subsection{Larger LLMs}\label{l-llm}

In the LLaMA3-8b and LLaMA2-13b experiments, we update certain labels in AMAN and TREC based on samples selected in LICL prior to the ICL experiments. In AMAN, `other’ is replaced with `noemotion’ in LLaMA3-8b and with `neutral’ in LLaMA2-13b. In TREC with LLaMA3-8b, `entity’ is replaced with `element’ `description’ with `definition,’ `human’ with `individual,’ and `location’ with `place’. In LLaMA2-13b, `entity’ is replaced with `word’, and `human’ with `persons’. Table \ref{LLMquality} shows the improved accuracy across models: AMAN and TREC experience increases of 3.8\% and 5.2\% respectively in the LLaMA3-8b model, while AMAN and TRED see increases of 1.28\% and 6.6\% respectively in the LLaMA2-13b model. These results demonstrate the superiority of our initial class-related word replacement method over the direct use of class names in enhancing 1-shot ICL performance, even with larger language models.

\begin{table}[htbp]
\centering
\resizebox{\linewidth}{!}{
\begin{tabular}{lcclcc}
\hline
\multicolumn{1}{c}{LICL}   & AMAN               & TREC              &  & AMAN               & TREC               \\ \hline
\textit{}                  & \multicolumn{2}{c}{\textit{LLaMA3-8b}} &  & \multicolumn{2}{c}{\textit{LLaMA2-13b}} \\
Original                   & 57.27              & 73.80             &  & 58.87              & 75.60              \\
\multicolumn{1}{r}{Update} & \textbf{61.15}     & \textbf{79.00}    &  & \textbf{60.15}     & \textbf{82.20}     \\ \hline
\end{tabular}
}
\caption{Enhancement in 1-shot ICL accuracy (\%) through initial class-related words updates in LLaMA3-8b and LLaMA2-13b.}
\label{LLMquality}
\end{table}

\section{Further Studies}\label{indepthanalysis}

In this section, we conducted more detailed experiments to analyze the effectiveness of our proposed approach for inserting multiple class-related words into demonstrations.

\subsection{Case Study: Effectiveness of Pool Refinement\label{sec:filtering}}
Table \ref{filteringinfo} lists the number of filtered words (-) for each dataset under two LLMs during point-biserial testing ($P_{r}$). Numerous words filtered indicate that although many words match the task topic definition at the linguistic level, they are not suitable as class-related words at the LLM level. This suggests that simple label-based voting, commonly used in many prompt-based methods, might harm LLM's ICL, as candidate words do not align with the task based on LLM's understanding.

\begin{table}[htbp]
\centering
\resizebox{\linewidth}{!}{
\begin{tabular}{cccccccc}
\hline
Model     & SST2 & CR   & IMDB & TREC & AMAN & ISEAR & AGNews \\ \hline
LLaMA2-7b & -316 & -237 & -240 & -32  & -78  & -78   & -1163  \\ \hline
GPT2-xl   & -397 & -375 & -389 & -42  & -68  & -118  & -648   \\ \hline
\end{tabular}
}
\caption{The statistic information of filtering results under two-stage filtering.}
\label{filteringinfo}
\end{table}

The point-biserial filtering method refines class-related word pool, influencing the selection of demonstration samples. We analyzed its effect by comparing samples selected with and without it. Findings indicate notable impacts on sample selection for the CR (in both LLaMA2-7b and GPT2-xl) and AMAN (in GPT2-xl) datasets. Table \ref{ablatiorpb} presents the comparative 1-shot ICL classification results, demonstrating that point-biserial filtering not only enhances sample selection but also boosts classification performance. Omitting this method reduces performance, confirming its effectiveness. Besides, even without this filtering, the addition of multiple class-related words still improves performance. 

\begin{table}[htbp]
\centering
\resizebox{\linewidth}{!}{
\begin{tabular}{rccc}
\hline
\multicolumn{1}{c}{LICL} & CR                 & CR               & AMAN             \\ \hline
\multicolumn{1}{c}{}     & \textit{LLaMA2-7b} & \textit{GPT2-xl} & \textit{GPT2-xl} \\
With rpb                 & 94.41              & 64.89            & 47.87            \\
+MLabels\_CN        & 95.15              & 69.95            & 49.62            \\ \hline
Without rpb              & 93.35              & 64.36            & 40.60             \\
+MLabels\_CN        & 94.68              & 65.43            & 48.12            \\ \hline
\end{tabular}}
\caption{1-shot ICL classification accuracy performance (\%) comparing scenarios with ( \lq\lq With rpb") and without (\lq\lq Without rpb") point-biserial filtering.}
\label{ablatiorpb}
\end{table}

\subsection{Case Study: Multiple Label Biases Evaluation}\label{sec: case1}

To address concerns regarding label biases with the insertion of multiple class-related words, we first evaluate the robustness by examining the standard deviation performance from baseline models using both single and multiple label insertions across five random sample iterations. Table \ref{stdvanilla} summarizes the results.

\begin{table}[htbp]
\centering
\resizebox{\linewidth}{!}{
\begin{tabular}{rlllllll}
\hline
\multicolumn{1}{c}{}                  & \multicolumn{1}{c}{SST2} & \multicolumn{1}{c}{CR} & \multicolumn{1}{c}{IMDB} & \multicolumn{1}{c}{TREC} & \multicolumn{1}{c}{AMAN} & \multicolumn{1}{c}{ISEAR} & \multicolumn{1}{c}{AGNews} \\ \hline
\multicolumn{1}{l}{\textit{LLaMA2-7b}} & \multicolumn{1}{c}{}     & \multicolumn{1}{c}{}   & \multicolumn{1}{c}{}     & \multicolumn{1}{c}{}     & \multicolumn{1}{c}{}     & \multicolumn{1}{c}{}      & \multicolumn{1}{c}{}       \\
Single Class-Related Word                     & 2.81e-4                  & 1.29e-4                & 8.77e-5                  & 2.01e-3                  & 9.97e-4                  & 5.38e-4                   & 9.69e-4                    \\
Multiple Class-Related Words                     & 1.37e-4                  & 8.70e-5                & 4.52e-5                  & 1.31e-3                  & 8.09e-4                  & 2.40e-4                   & 9.25e-5                    \\ \hline
\multicolumn{1}{l}{\textit{GPT2-xl}}   & \multicolumn{1}{c}{}     & \multicolumn{1}{c}{}   & \multicolumn{1}{c}{}     & \multicolumn{1}{c}{}     & \multicolumn{1}{c}{}     & \multicolumn{1}{c}{}      & \multicolumn{1}{c}{}       \\
Single Class-Related Word                        & 1.03e-2                  & 1.63e-05               & 1.34e-2                  & 2.23e-03                 & 5.15e-3                  & 8.17e-4                   & 2.68e-3                    \\
Multiple Class-Related Words                       & 1.47e-3                  & 5.66e-06               & 9.03e-3                  & 1.24e-03                 & 4.84e-3                  & 7.20e-4                   & 1.55e-3                    \\ \hline
\end{tabular}}
\caption{Standard Deviation Evaluation: Results from experiments in vanilla and vanilla + Demo-MLabels\_LW configurations.}
\label{stdvanilla}
\end{table}

Our findings suggest that using multiple class-related words in our demonstrations does not introduce new label biases, but rather enhances robustness, as indicated by a lower standard deviation compared to the single label approach. Additionally, as shown in Table \ref{LICLoverperformance}, the classification performance has improved under both single and multiple class-related words scenarios compared to the baseline models.

We also conducted an additional experiment for comparison with Domain-context calibration (DC) \cite{fei2023mitigating}, a label bias calibration method. In this experiment, the DC model is applied to GPT2-xl across three datasets: SST2, TREC, and AGNews. The training and testing conditions are the same as our approach to ensure a fair comparison. The in-context examples in DC are 1-shot and label-balanced, similar to our setup. Other settings, such as the number of seeds, follow the default settings\footnote{\url{https://github.com/fywalter/label-bias}}.

Table \ref{Dc} details the accuracy performance for DC and our method (LICL). Our method outperforms DC across the evaluated datasets, highlighting its effectiveness in improving classification accuracy in varied contexts without increasing label bias.

\begin{table}[htbp]
\centering
\resizebox{0.9\linewidth}{!}{
\begin{tabular}{lccc}
\hline
\textit{GPT2-xl}                                              & SST2           & TREC           & AGNews         \\ \hline
DC                                                            & 85.60          & 58.90          & 76.90          \\
\cellcolor[HTML]{ECF4FF}LICL                                  & 85.17          & 70.00          & 78.92          \\
\multicolumn{1}{r}{\cellcolor[HTML]{ECF4FF}+MLabels\_CN} & 91.65          & 70.40          & 79.08          \\
\multicolumn{1}{r}{\cellcolor[HTML]{ECF4FF}+MLabels\_LW} & \textbf{91.65} & \textbf{70.40} & \textbf{79.49} \\ \hline
\end{tabular}}
\caption{Accuracy Performance (\%) Comparison between DC and LICL on GPT2-xl.}
\label{Dc}
\end{table}

\subsection{Case Study: Analysis of Utilizing Multiple Class-Related Words Mapping in Prediction 
 \label{sec:demonstrationprediction}}
Table \ref{LICLoverperformance} and \ref{5shot} show that `+MLabels\_LW' (prediction based on the maximum logit over inserted class-related words) achieves higher accuracy than `+MLabels\_CN' (prediction based on the maximum logit over class names) in some datasets. This seems to align with the idea that incorporating extra class-related word mapping in the final prediction can enhance prompt-based methods \cite{schick2021exploiting}. However, in ICL, models tend to predict based on the labels provided in demonstrations, which may perform differently with extra class-related word mappings in prediction. We compare the effectiveness of using extra class-related word mappings only in predictions versus including them in demonstrations for both binary-class and multi-class tasks. Additionally, we evaluate the impact of leveraging extra class-related word mappings under zero-shot settings.
\begin{table}[htbp]
\centering
\resizebox{\linewidth}{!}{
\begin{tabular}{rcccccc}
\hline
\multicolumn{1}{c}{}                   & SST2  & TREC  & AMAN  & ISEAR & AGNews & Avg.           \\ \hline
\multicolumn{1}{l}{\textit{LLaMA2-7b}} &       &       &       &       &        & \multicolumn{1}{l}{} \\
\multicolumn{1}{l}{ZSL}                & 88.96 & 68.80 & 48.87 & 58.60 & 67.29  & 66.50                \\
\cellcolor[HTML]{ECF4FF}ZSL\_$P_{r}$        & 90.94 & 60.00$\downarrow$ & 47.37$\downarrow$ & 60.80 & 60.03$\downarrow$  & 63.83                \\
\multicolumn{1}{l}{LICL}               & 95.39 & 78.40 & 59.90 & 72.49 & 78.06  & 76.85                \\
\cellcolor[HTML]{ECF4FF}LICL\_$P_{r}$       & 95.44 & 72.00$\downarrow$ & 63.13 & 71.56$\downarrow$ & 74.37$\downarrow$  & 75.30                \\
+MLabels\_CN                          & 95.97 & 79.80 & 65.16 & 73.55 & 86.55  & 80.21                \\
+MLabels\_LW                         & 95.97 & 80.60 & 69.40 & 73.09 & 86.58  & 81.13                \\
\cellcolor[HTML]{ECF4FF}+MLabels\_$P_{r}$  & 94.51$\downarrow$ & 74.40$\downarrow$ & 67.64$\downarrow$ & 71.56$\downarrow$ & 75.33$\downarrow$  & 76.69                \\ \hline
\multicolumn{1}{l}{\textit{GPT2-xl}}   &       &       &       &       &        & \multicolumn{1}{l}{} \\
\multicolumn{1}{l}{ZSL}                & 79.57 & 38.00 & 39.60 & 42.33 & 53.16  & 50.53                \\
\cellcolor[HTML]{ECF4FF}ZSL\_$P_{r}$        & 50.30$\downarrow$ & 46.80 & 37.84$\downarrow$ & 42.52 & 51.29$\downarrow$  & 45.75                \\
\multicolumn{1}{l}{LICL\_C}               & 85.17 & 61.60 & 47.87 & 48.64 & 78.92  & 64.44                \\
\cellcolor[HTML]{ECF4FF}LICL\_$P_{r}$       & 85.17 & 67.60 & 47.62$\downarrow$ & 48.70 & 65.28$\downarrow$  & 62.87                \\
+MLabels\_CN                          & 91.65 & 70.40 & 49.62 & 59.73 & 79.08  & 70.10                \\
+MLabels\_LW                         & 91.65 & 70.40 & 49.87 & 58.74 & 79.49  & 70.03                \\
\cellcolor[HTML]{ECF4FF}+MLabels\_$P_{r}$  & 91.65 & 68.80$\downarrow$ & 49.87 & 58.64$\downarrow$ & 67.58$\downarrow$  & 67.31                \\ \hline
\end{tabular}
}
\caption{Impact of prediction under class names, multiple class-related words in demonstration and $P_{r}$. Arrow $\downarrow$ indicates a decrease in accuracy of predictions over $P_{r}$, compared to those over class names (CN) or inserted class-related words (LW).}
\label{prediction}
\end{table}

In Table \ref{prediction}, most experiments show decreased performance when predicting based on class-related words in $P_{r}$ compared to predictions on class-related words that appeared in demonstrations (including class names). This suggests that simply applying extra class-related word mappings in the final prediction might disrupt information learned from demonstrations in ICL. Surprisingly, even under zero-shot learning (ZSL), where label information isn't prompted, adding extra label knowledge still decreases some classification performance. These results highlight the complex role of labels in ICL classification.

\subsection{Case Study: Effectiveness of Label Balance in LICL\label{sec:labelbalance}}
Our method is evaluated under a label-balanced demonstration setting, assuming that every category of label information matters. We also investigate an unbalanced setting by removing a sample-label pair with the highest sample score according to our scoring methods. Table \ref{unbalance} summarizes the results for two language models in 1-shot ICL.

\begin{table}[htbp]
\centering
\resizebox{\linewidth}{!}{
\begin{tabular}{rcccccccc}
\hline
\multicolumn{1}{c}{}                                                                               & SST2           & CR             & IMDB           & TREC           & AMAN           & ISEAR          & AGNews         & Avg.     \\ \hline
\multicolumn{1}{l}{\textit{LLaMA2-7b}}                                                             &                &                &                &                &                &                &                &                \\
\multicolumn{1}{l}{LICL}                                                                           & 95.39          & 94.41          & 95.40          & 78.40          & 59.90          & 72.49          & 84.11          & 82.87          \\
\cellcolor[HTML]{ECF4FF}\textbf{\begin{tabular}[c]{@{}r@{}}LICL\\ unbalanced\end{tabular}}         & 91.98          & 93.35          & 93.60          & 68.00          & 57.39          & 70.70          & 81.49          & 79.50          \\
+MLabels\_CN                                                                                      & 95.97          & 95.15          & 95.60          & 79.80          & 65.16          & \textbf{73.55} & 86.55          & 84.54          \\
\cellcolor[HTML]{ECF4FF}\textbf{\begin{tabular}[c]{@{}r@{}}+MLabels\_CN\\ unbalanced\end{tabular}} & 92.42          & 93.62          & 94.40          & 69.60          & 62.16          & 71.36          & 85.87          & 81.35          \\
+MLabels\_LW                                                                                     & \textbf{95.97} & \textbf{95.15} & \textbf{95.60} & 80.60          & \textbf{69.40} & 73.09          & \textbf{86.58} & \textbf{85.20} \\
\cellcolor[HTML]{ECF4FF}\textbf{\begin{tabular}[c]{@{}r@{}}+MLabels\_LW\\ unbalanced\end{tabular}} & 92.42          & 93.62          & 94.40          & 81.80          & 62.91          & 71.36          & 86.36          & 83.27          \\ \hline
\multicolumn{1}{l}{\textit{GPT2-xl}}                                                               &                &                &                &                &                &                &                &                \\
\multicolumn{1}{l}{LICL}                                                                           & 85.17          & 64.89          & 71.50          & 70.00          & 47.87          & 48.64          & 78.92          & 66.71          \\
\cellcolor[HTML]{ECF4FF}\textbf{\begin{tabular}[c]{@{}r@{}}LICL\\ unbalanced\end{tabular}}         & 51.02          & 73.67 & 67.80          & 53.60          & 41.60          & 54.62          & 34.45          & 53.82          \\
+MLabels\_CN                                                                                      & 91.65          & 69.95          & 73.40          & 70.40          & 49.62          & \textbf{59.73} & 79.08          & \textbf{70.55} \\
\cellcolor[HTML]{ECF4FF}\textbf{\begin{tabular}[c]{@{}r@{}}+MLabels\_CN\\ unbalanced\end{tabular}} & 52.85          & 74.47          & 68.20          & 55.40          & 45.86          & 56.75          & 43.75          & 56.75          \\
+MLabels\_LW                                                                                     & \textbf{91.65} & 69.95          & \textbf{73.40} & \textbf{70.40} & \textbf{49.87} & 58.74          & \textbf{79.49} & 70.50          \\
\cellcolor[HTML]{ECF4FF}\textbf{\begin{tabular}[c]{@{}r@{}}+MLabels\_LW\\ unbalanced\end{tabular}} & 52.85          & \textbf{74.47}          & 68.40          & 59.60          & 45.61          & 56.42          & 43.75          & 57.30          \\ \hline
\end{tabular}
}
\caption{Accuracy performance (\%) of LICL, `+MLabels\_CN', `+MLabels\_LW' under label-balanced and label-unbalanced demonstrations.} 
\label{unbalance}
\end{table}

The impact of label-unbalanced demonstrations varies between the two models. For LLaMA2-7b, a large-size language model, the negative effects of missing label information are less marked, with an average accuracy decrease of 3.37\%. The most significant accuracy drops in TREC by 10.40\%. The leverage of multiple class-related words mitigated the performance loss, reducing the average accuracy decline to 1.93\% in `+MLabels\_LW' compared to a label-balanced setting.
In contrast, GPT2-xl, a smaller model with 1.5 billion parameters, experienced a marked performance decline under unbalanced conditions, averaging a 12.89\% decrease in accuracy. Specifically, SST2 and AGNews experienced over 35\% declines. Interestingly, CR demonstrated improved performance despite the unbalanced labels. The incorporation of multiple class-related words consistently enhanced performance across all datasets in the unbalanced setting, affirming their utility in ICL.

\subsection{Case Study: Sample-Label Logit Separability Visualization}\label{sec:visualization}

\begin{figure*}[htbp]
\centering  
\subfigure[Logit separability for selected positive sample]{
\label{good positive sample}
\includegraphics[width=0.48\textwidth]{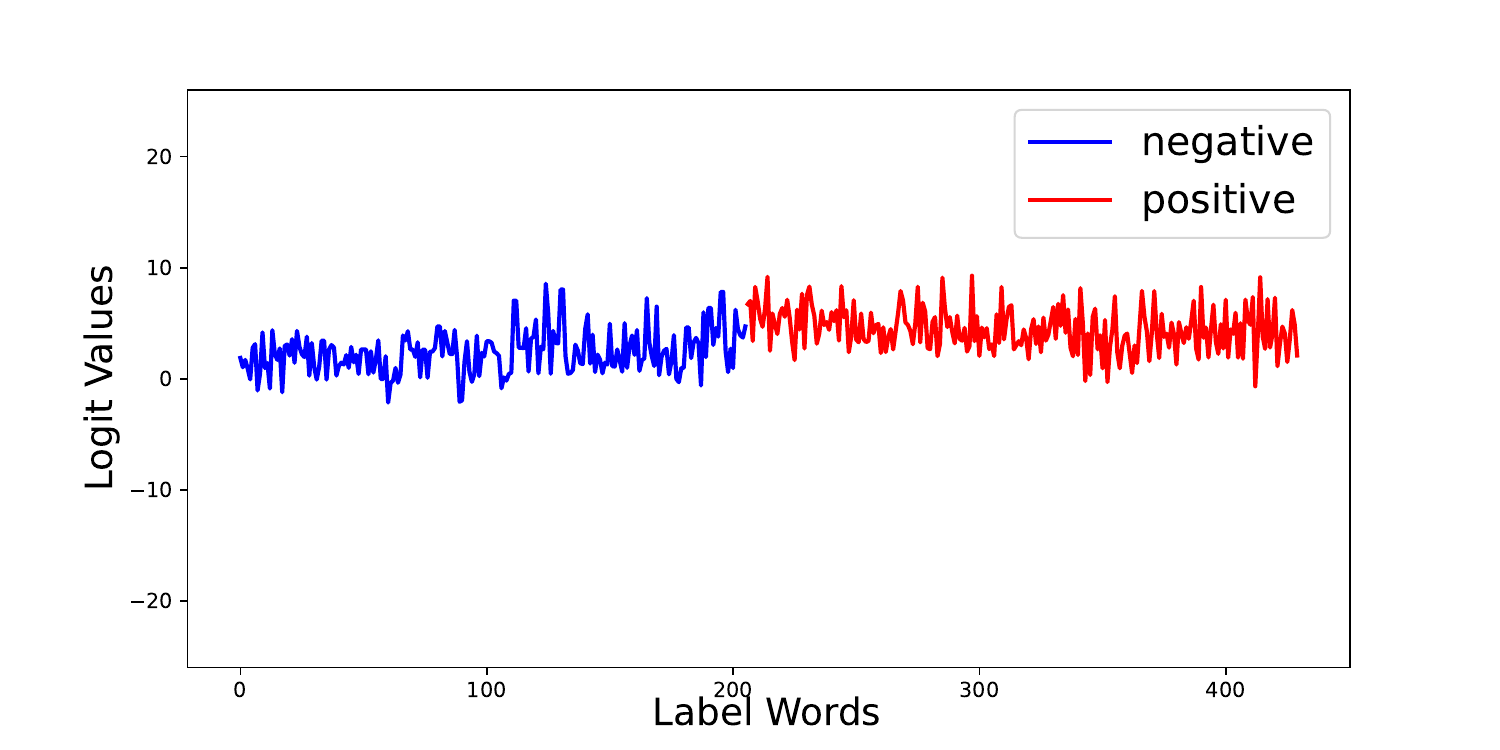}}
\subfigure[Logit separability for selected negative sample]{
\label{good negative sample}
\includegraphics[width=0.48\textwidth]{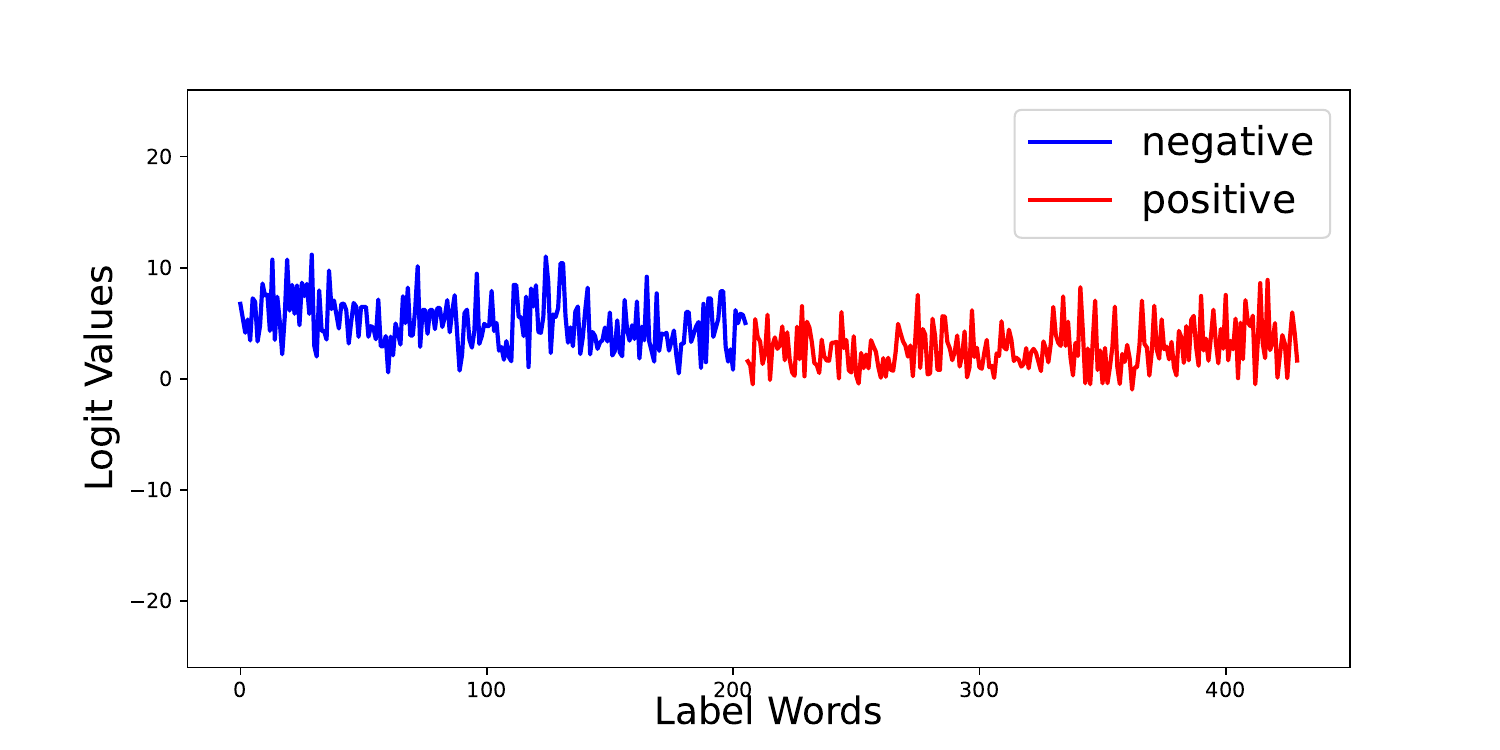}}
\subfigure[Logit separability for a positive sample]{
\label{bad positive sample}
\includegraphics[width=0.48\textwidth]{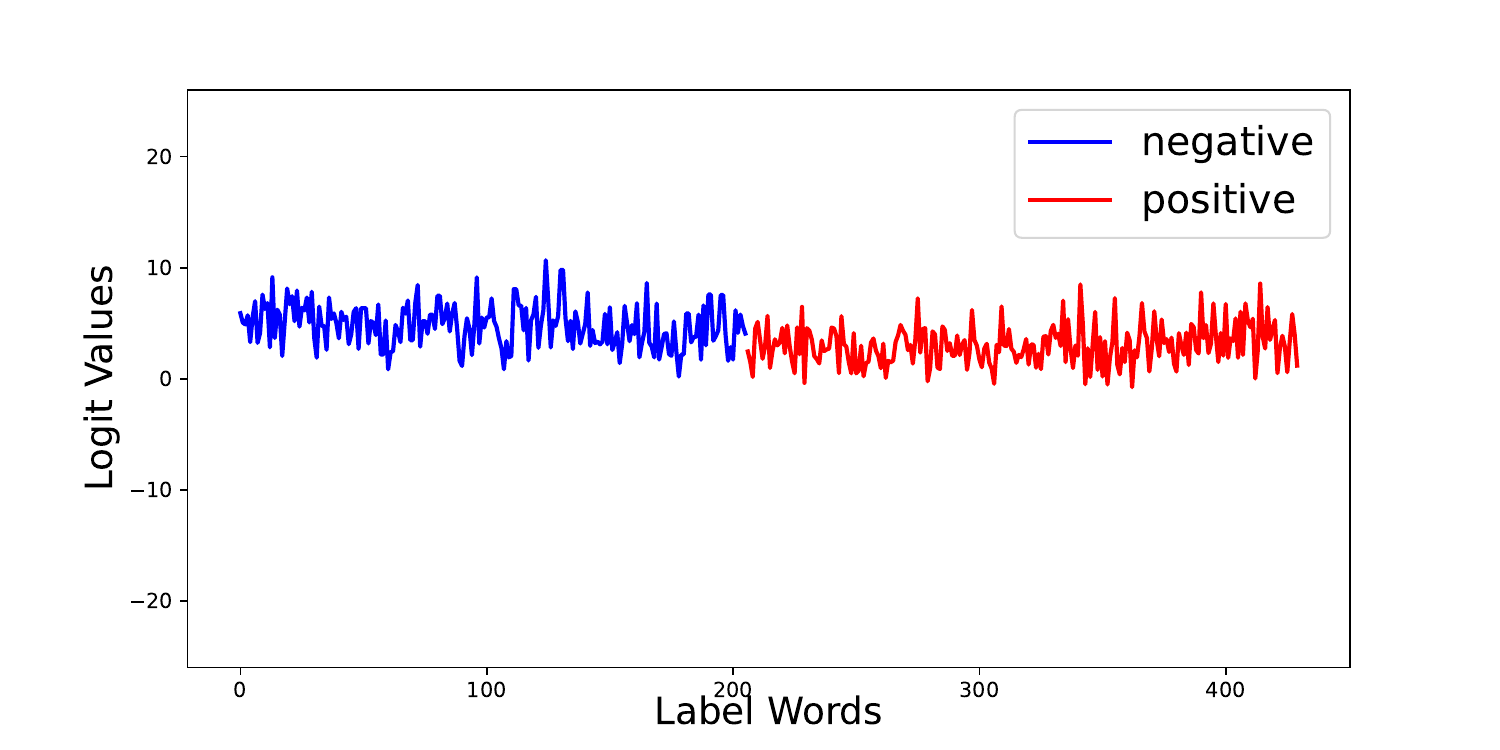}}
\subfigure[Logit separability for a negative sample]{
\label{bad negative sample}
\includegraphics[width=0.48\textwidth]{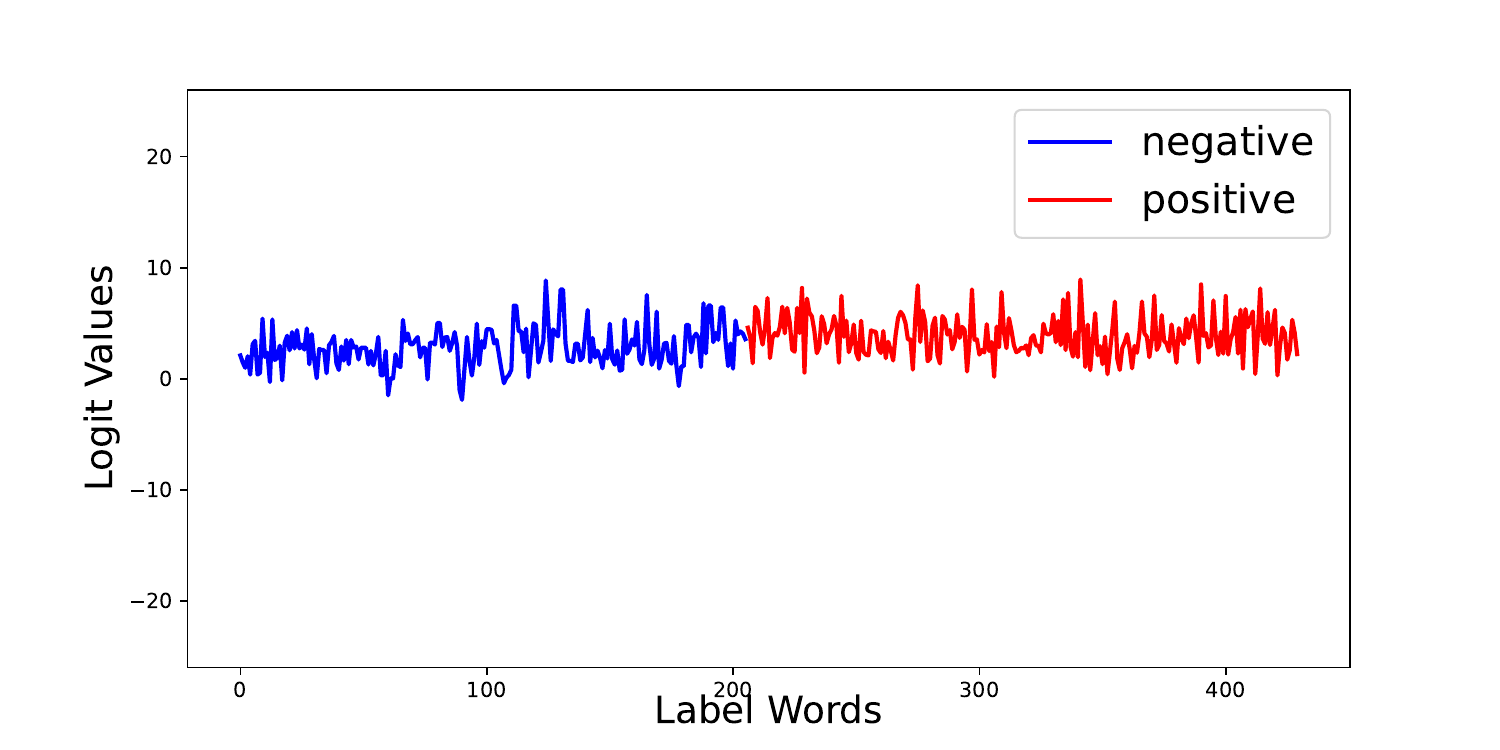}}
\caption{Visualization of samples' logit separability across class-related words in LLaMA3-8b, evaluated on the SST2 training set using a refined verbalizer. The class-related words are categorized into negative (0-205) and positive (206-429) groups.}
\label{sample-label}
\end{figure*}

\begin{figure*}[htbp]
\centering  
\subfigure[Logit separability for selected positive word]{
\label{good positive word}
\includegraphics[width=0.48\textwidth]{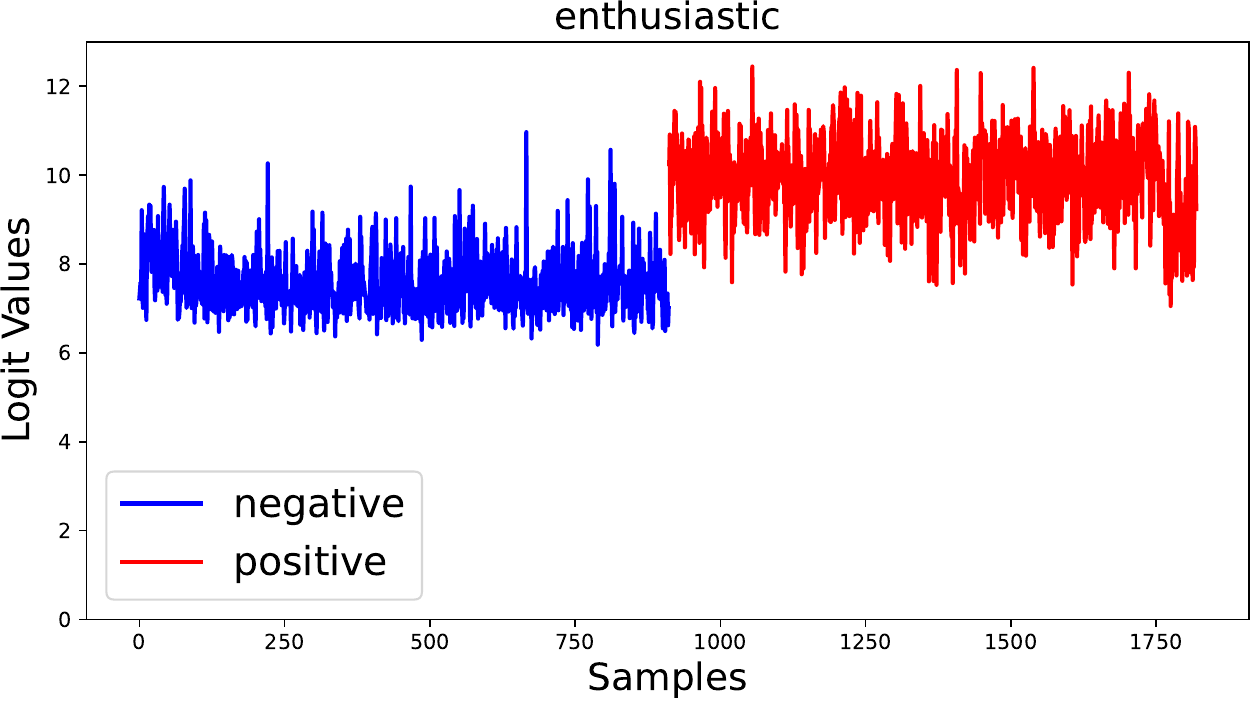}}
\subfigure[Logit separability for selected negative word]{
\label{good negative word}
\includegraphics[width=0.48\textwidth]{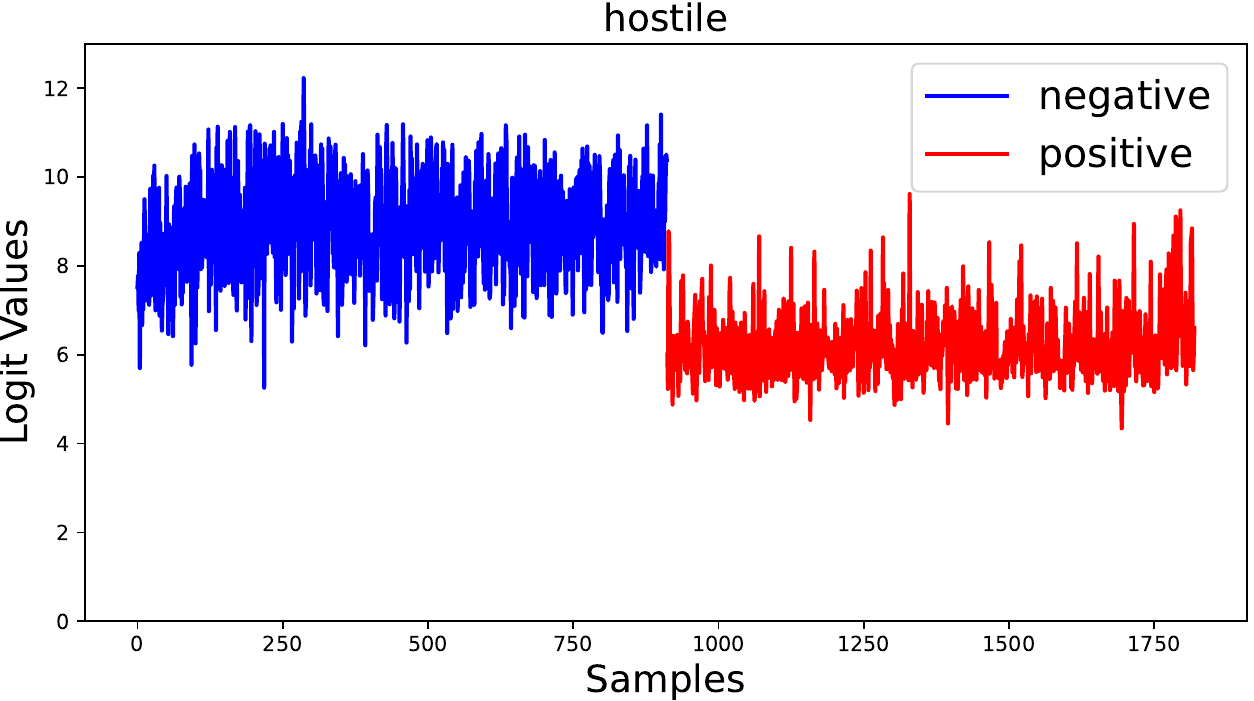}}
\subfigure[Logit separability for a positive word]{
\label{bad positive word}
\includegraphics[width=0.48\textwidth]{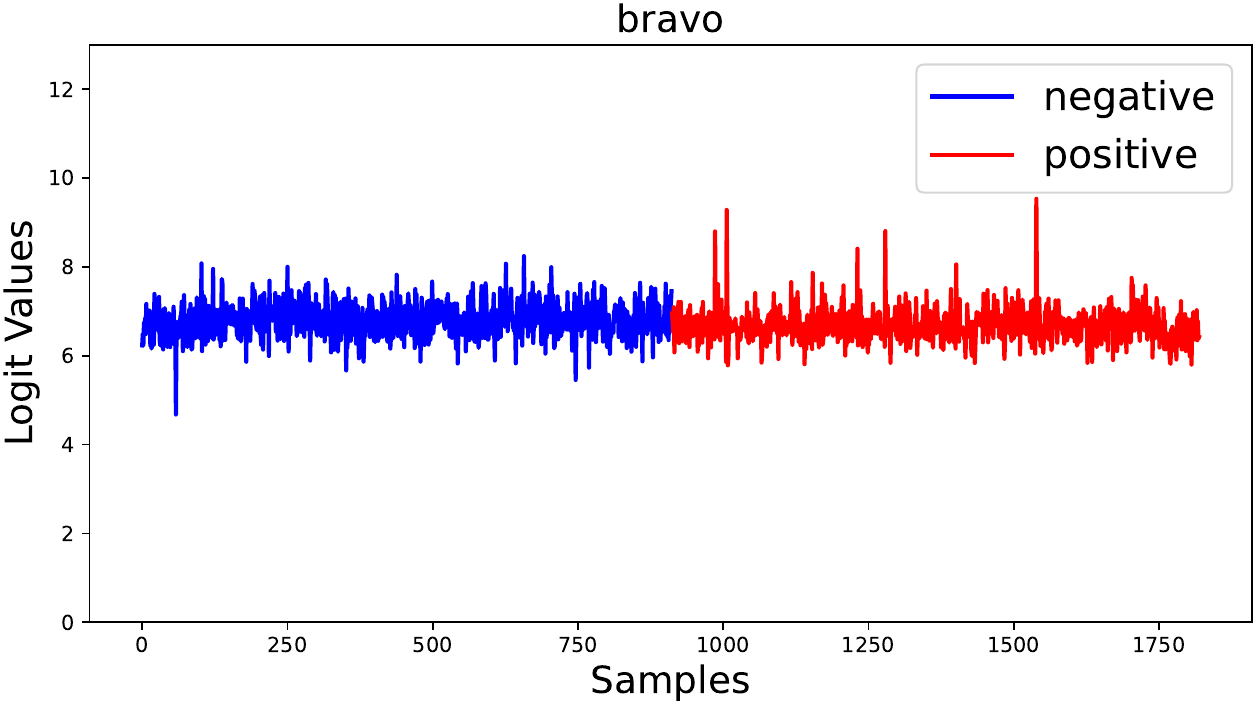}}
\subfigure[Logit separability for a negative word]{
\label{bad negative word}
\includegraphics[width=0.48\textwidth]{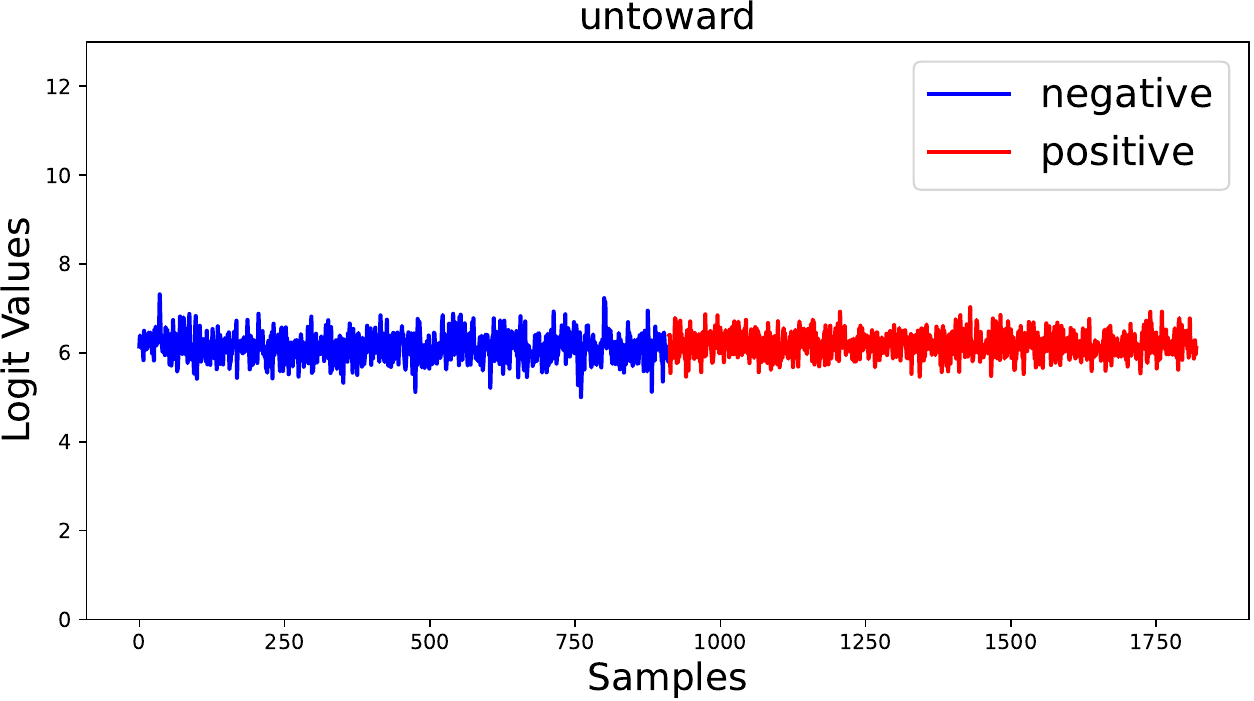}}
\caption{Visualization of class-related words' logit separability in LLaMA3-8b, evaluated on the SST2 testing set using a refined verbalizer. The samples are categorized into negative (0-911) and positive (912-1820) groups.}
\label{label-sample}
\end{figure*}

To assess the effectiveness of sample selection based on the discriminative logit separability of class-related words and the insertion of class-related words into test samples, Fig.\ref{sample-label} illustrates the distribution of class-related word logit for selected and other samples, while Fig.\ref{label-sample} displays the logit separability of inserted class-related word logit under 1-shot ICL learning across testing samples, along with other lass-related word logit separability under the same conditions. 

Comparing Fig.\ref{good positive sample} with Fig.\ref{bad positive sample}, and Fig.\ref{good negative sample} with Fig.\ref{bad negative sample}, it is evident that the selected samples demonstrate superior logit separability between corresponding polarity class-related words and those of the opposite polarity, which enhances class prediction accuracy.

Similarly, an analysis of Fig.\ref{good positive word} against Fig.\ref{bad positive word}, and Fig.\ref{good negative word} against Fig.\ref{bad negative word}, shows that selected words exhibit improved logit separability between samples of matching polarity and those of opposing polarity, contributing positively to class prediction.

\subsection{Case Study: Inference Cost Evaluation} \label{sec: case4}

To address concerns about the inference costs associated with the insertion of additional multiple class-related words, we evaluated the inference time of all compared methods and ours in 1-shot ICL experiments. The primary factors affecting inference costs are the length of the demonstration sample and the size of the language model. Since all demonstrations are 1-shot and balanced in terms of label distribution, variations in sample length directly impact inference times. Adding multiple class-related words increases the length of label inputs beyond using just the class name.

Table \ref{inferencetime} details the running times for test sets using both our method and comparative models, and includes the incremental times associated with increasing the number of class-related words (in Figs. \ref{validationN} and \ref{numberofword}) in demonstrations, marked as `+MLabels time increased'.

\begin{table*}[htbp]
\centering
\resizebox{\linewidth}{!}{
\begin{tabular}{lccccccclccccccc}
\hline
\textit{Llama2-7b}                                                                        & SST2   & CR    & IMDB   & TREC  & AMAN  & ISEAR  & AGNews   & \textit{GPT2-xl}                                                                          & SST2  & CR  & IMDB  & TREC & AMAN & ISEAR & AGNews \\
Topk                                                                                      & 10m46s & 2m35s & 55m27s & 2m39s & 3m46s & 21m48s & 1h40m32s & Topk                                                                                      & 1m53s & 22s & 6m8s  & 31s  & 50s  & 5m55s & 24m30s \\
SelfICL                                                                                   & 11m9s  & 2m19s & 49m46s & 3m    & 3m45s & 19m11s & 1h44m37s & SelfICL                                                                                   & 2m    & 24s & 6m18s & 32s  & 48s  & 4m49s & 24m55s \\
DataICL                                                                                   & 7m5s   & 1m58s & 52m32s & 2m37s & 3m4s  & 15m51s & 1h34m37s & DataICL                                                                                   & 1m56s & 24s & 6m13s & 31s  & 43s  & 6m25s & 18m46s \\
LICL                                                                                      & 9m41s  & 2m12s & 38m52s & 2m31s & 3m22s & 15m30s & 1h29m54s & LICL                                                                                      & 1m26s & 22s & 6m20s & 21s  & 41s  & 4m45s & 19m7s  \\
\multicolumn{1}{r}{\begin{tabular}[c]{@{}r@{}}+MLabels\\ time increased\end{tabular}} & +2s    & +6s   & +58s   & +3s   & +16s  & +18s   & +215s    & \multicolumn{1}{r}{\begin{tabular}[c]{@{}r@{}}+MLabels\\ time increased\end{tabular}} & +1s   & +1s & +7s   & +2s  & +1s  & +10s  & +26s   \\ \hline
\end{tabular}}
\caption{Inference Time Evaluation (where `h' denotes hours, `m' denotes minutes, and `s' denotes seconds).}
\label{inferencetime}
\end{table*}

Under 1-shot demonstration conditions, the inference times for our method are comparable to, or even faster than, those of other methods, given the length of the selected samples. With the addition of multiple labels, the inference time increases by minutes or seconds, which we consider to be within an acceptable range, given the enhanced performance and accuracy achieved.

\end{document}